\documentclass[lettersize,journal]{IEEEtran}
\IEEEoverridecommandlockouts                              

\usepackage{00_ps}

\preprinttrue
\publishedtrue

\renewcommand{\copyrightOwnerFull}{IEEE}
\renewcommand{\copyrightOwnerShort}{IEEE}

\glsdisablehyper 
\DeclareAcronym{adas}{
    short = ADAS,
    long  = Action Propagation,
}
\DeclareAcronym{ap}{
    short = AP,
    long  = Action Propagation,
}
\DeclareAcronym{api}{
    short = API,
    long  = Application Programming Interface,
    short-indefinite = an,
}
\DeclareAcronym{c2c}{
    short = C2C,
    long  = Center-to-Center,
}
\DeclareAcronym{cav}{
    short = CAV,
    long  = Connected and Automated Vehicle,
}
\DeclareAcronym{cbf}{
    short = CBF,
    long  = Control Barrier Function,
}
\DeclareAcronym{cg}{
    short = CG,
    long = Center of Gravity,
    short-plural = s,
    long-plural-form = Centers of Gravity,
}
\DeclareAcronym{cnn}{
    short = CNN,
    long  = Convolutional Neural Network
}
\DeclareAcronym{cpm}{
    short = CPM,
    long  = Cyber-Physical Mobility
}
\DeclareAcronym{cpmlab}{
    short = CPM Lab,
    long  = Cyber-Physical Mobility Lab
}
\DeclareAcronym{dmpc}{
    short = DMPC,
    long  = distributed model predictive control
}
\DeclareAcronym{dql}{
    short = DQL,
    long  = Deep Q-Learning
}

\DeclareAcronym{its}{
    short = ITS,
    long  = Intelligent Transportation Systems,
    short-indefinite = an,
}
\DeclareAcronym{mappo}{
    short = MAPPO,
    long  = Multi-Agent \ac{ppo},
    short-indefinite = an,
}
\DeclareAcronym{maddpg}{
    short = MADDPG,
    long  = Multi-Agent Deep Deterministic Policy Gradient,
    short-indefinite = an,
}
\DeclareAcronym{mas}{
    short = MAS,
    long  = Multi-Agent System,
    short-indefinite = an,
}
\DeclareAcronym{mdp}{
    short = MDP,
    long  = Markov decision process,
    short-indefinite = an,
}
\DeclareAcronym{mg}{
    short = MG,
    long  = Markov Game,
    short-indefinite = an,
}
\DeclareAcronym{ml}{
    short = ML,
    long  = Machine Learning,
    short-indefinite = an,
}
\DeclareAcronym{mtv}{
    short = MTV,
    long  = Minimum Translation Vector,
    short-indefinite = an,
}
\DeclareAcronym{mpc}{
    short = MPC,
    long  = model predictive control,
    short-indefinite = an,
}
\DeclareAcronym{marl}{
    short = MARL,
    long  = Multi-Agent Reinforcement Learning,
    short-indefinite = an,
}
\DeclareAcronym{ocp}{
    short = OCP,
    long  = Optimal Control Problem,
    short-indefinite = an,
    long-indefinite = an,
}
\DeclareAcronym{per}{
    short = PER,
    long  = Prioritized Experience Replay
}
\DeclareAcronym{pomdp}{
    short = POMDP,
    long  = Partially Observable \ac{mdp}
}
\DeclareAcronym{pomg}{
    short = POMG,
    long  = Partially Observable \ac{mg}
}
\DeclareAcronym{ppo}{
    short = PPO,
    long  = Proximal Policy Optimization
}
\DeclareAcronym{ros}{
    short = ROS,
    long  = Robot Operating System,
    short-indefinite = a,
}
\DeclareAcronym{rhc}{
    short = RHC,
    long  = receding horizon control,
    short-indefinite = an,
}
\DeclareAcronym{rl}{
    short = RL,
    long  = Reinforcement Learning,
    short-indefinite = a,
}
\DeclareAcronym{sat}{
    short = SAT,
    long = Separating Axis Theorem,
    short-indefinite = an
}
\DeclareAcronym{sim2real}{
    short = Sim2Real,
    long  = Simulation-To-Reality,
    short-indefinite = a,
}
\DeclareAcronym{som}{
    short = SOM,
    long  = Self Other-Modeling,
    short-indefinite = a,
}
\DeclareAcronym{zsg}{
    short = ZSG,
    long  = Zero-Shot Generalization,
}
\input{00_copyright.tex}

\begin{document}
\title{
    Small-Scale Testbeds for Connected and Automated Vehicles and Robot Swarms: Challenges and a Roadmap
}

\author{
    Jianye Xu$^{1}$\,\orcidlink{0009-0001-0150-2147},
    Johannes Betz$^{3}$\,\orcidlink{0000-0001-9197-2849},
    Armin Mokhtarian$^{1}$\,\orcidlink{0000-0002-5345-4538},
    Archak Mittal$^{4}$\,\orcidlink{0000-0001-6186-4513},
    Mengchi Cai$^{5}$\,\orcidlink{0000-0002-8681-7067},
    Rahul Mangharam$^{6}$\,\orcidlink{0000-0002-3388-8283},
    Omar M. Shehata$^{7}$\,\orcidlink{0000-0002-3604-3534},
    Catherine M. Elias$^{8}$\,\orcidlink{0000-0002-1444-9816},
    Jan-Nico Zaech$^{9}$\,\orcidlink{0000-0003-2566-0841},
    Patrick Scheffe$^{1,12}$\,\orcidlink{0000-0002-2707-198X},
    Felix Jahncke$^{3}$\,\orcidlink{0009-0003-4484-8361},
    Sangeet Sankaramangalam Ulhas$^{10}$\,\orcidlink{0009-0006-1292-8959},
    Kaj Munhoz Arfvidsson$^{11}$\,\orcidlink{0009-0007-3871-5828},
    Bassam Alrifaee$^{2}$\,\orcidlink{0000-0002-5982-021X}
    \thanks{$^{1}$Department of Computer Science, RWTH Aachen University, Germany, \texttt{\{xu, mokhtarian, scheffe\}@embedded.rwth-aachen.de}}
    \thanks{$^{2}$Department of Aerospace Engineering, University of the Bundeswehr Munich, Germany, \texttt{bassam.alrifaee@unibw.de}}
    \thanks{$^{3}$Professorship of Autonomous Vehicle Systems, Technical University of Munich, Germany, \texttt{\{johannes.betz, felix.jahncke\}@tum.de}}
    \thanks{$^{4}$Department of Civil Engineering, Indian Institute of Technology Bombay, India, \texttt{archak@civil.iitb.ac.in}}
    \thanks{$^{5}$School of Vehicle and Mobility, Tsinghua University, China, \texttt{caimengchi@tsinghua.edu.cn}}
    \thanks{$^{6}$Department of Electrical and Systems Engineering, University of Pennsylvania, USA, \texttt{rahulm@seas.upenn.edu}}
    \thanks{$^{7}$Mechatronics Engineering Department, Faculty of Engineering and Materials Science, the German University in Cairo, Egypt, \texttt{omar.mohamad@guc.edu.eg}}
    \thanks{$^{8}$Department of Computer Science Engineering, the German University in Cairo, \texttt{catherine.elias@ieee.org}}
    \thanks{$^{9}$INSAIT, Sofia University “St. Kliment Ohridski”, Bulgaria, \texttt{jan-nico.zaech@insait.ai}}
    \thanks{$^{10}$School for Engineering of Matter, Transport and Energy, Arizona State University, USA, \texttt{sulhas@asu.edu}}
    \thanks{$^{11}$Division of Decision and Control Systems, KTH Royal Institute of Technology, Sweden, \texttt{kajarf@kth.se}}
    \thanks{$^{12}$Department of Mechanical Engineering, KU Leuven, Belgium, \texttt{patrick.scheffe@kuleuven.be}, and Flanders Make@KU Leuven, Belgium}
}
    \maketitle
\pagestyle{plain}
\begin{abstract}
This article proposes a roadmap to address the current challenges in small-scale testbeds for \acp{cav} and robot swarms. The roadmap is a joint effort of participants in the workshop ``1\textsuperscript{st} Workshop on Small-Scale Testbeds for Connected and Automated Vehicles and Robot Swarms,'' held on June 2 at the IEEE Intelligent Vehicles Symposium (IV) 2024 in Jeju, South Korea. The roadmap contains three parts:
\begin{enumerate*}
    \item enhancing accessibility and diversity, especially for underrepresented communities,
    \item sharing best practices for the development and maintenance of testbeds, and
    \item global testbed connectivity, e.g., through an abstraction layer, to support collaboration.
\end{enumerate*}
The workshop features eight invited speakers, four contributed papers \cite{lyons2024dart, ulhas2024ganbased, arfvidsson2024smallscale, hamza2024visionbased}, and a presentation of a survey paper on testbeds \cite{mokhtarian2024survey}. The survey paper provides an online comparative table of more than 25 testbeds, available at \href{https://bassamlab.github.io/testbeds-survey}{https://bassamlab.github.io/testbeds-survey}. The workshop's own website is available at \href{https://cpm.lrt.unibw.de/workshop}{https://cpm.lrt.unibw.de/workshop}.
\par\medskip
\end{abstract}


\acresetall  

\section{Introduction}\label{sec:introduction}
\IEEEPARstart{T}{he} advancement of \acp{cav} and robot swarms requires robust platforms for testing and validation. Small-scale testbeds offer a cost-effective, flexible, and scalable solution in controlled settings, making them critical for developing new technologies. Despite their potential, several challenges remain. We classify these challenges into two categories. The first category contains \textbf{transition challenges}, which are specific to small-scale testbeds. We focus on two transition challenges, i.e., transition from simulation to reality (also known as \ac{sim2real} gap) and transition from small- to full-scale. The second category includes challenges in small-scale testbeds that are also present in the real world, thus termed \textbf{inherent real-world challenges}. Examples are environmental unpredictability and distributed computing demands. 

\begin{figure*}[t]
    \centering
    \includegraphics[width=1\textwidth]{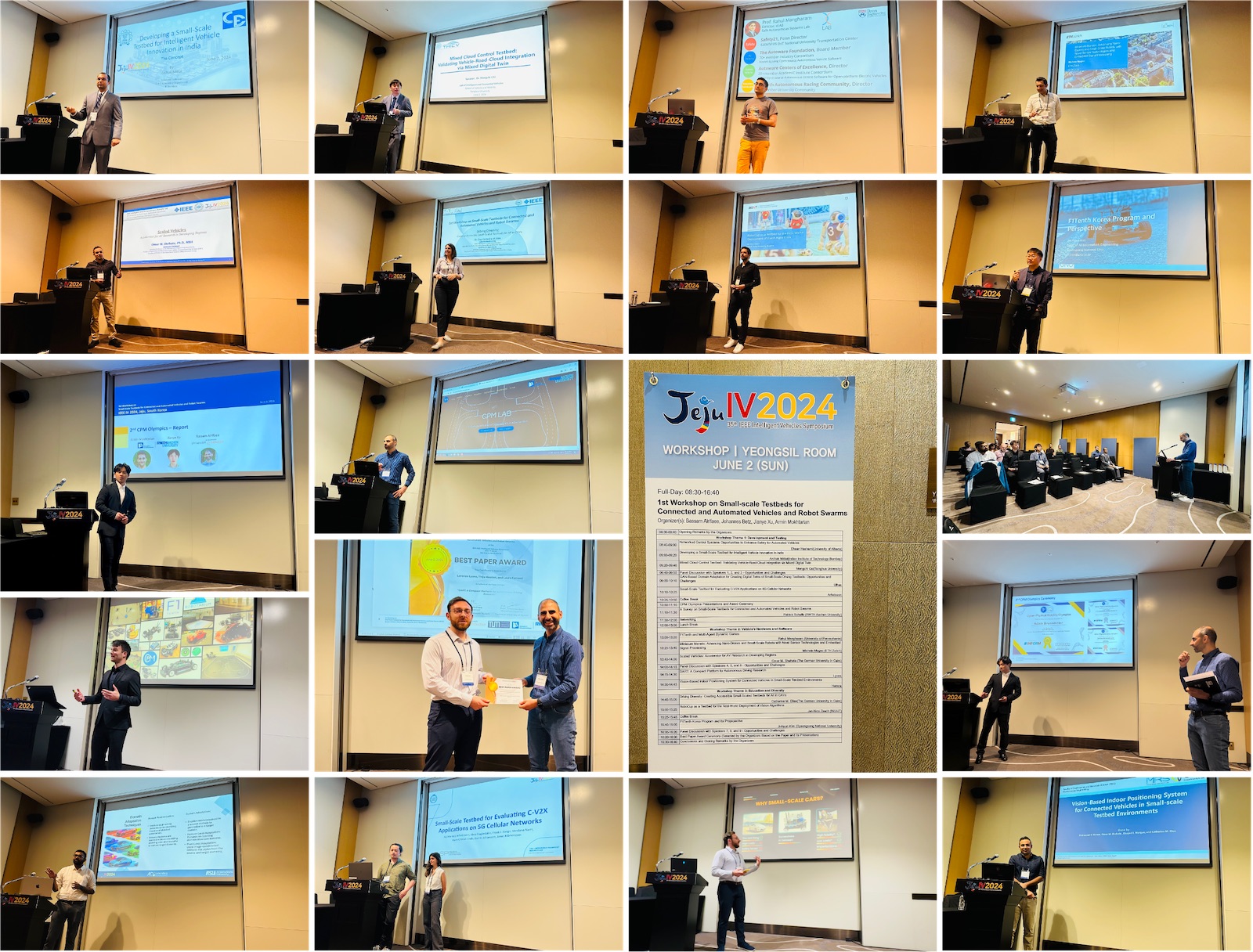}
    \caption{A collage of photographs from the workshop.}
    \label{fig_collage}
\end{figure*}

To address these challenges, we organized the \textit{1\textsuperscript{st} Workshop on Small-Scale Testbeds for Connected and Automated Vehicles and Robot Swarms} on June 2 at the IEEE Intelligent Vehicles Symposium (IV) 2024 in Jeju, South Korea. Our workshop brought together experts, researchers, and educators to discuss these challenges. The workshop's website is available at \href{https://cpm.lrt.unibw.de/workshop}{https://cpm.lrt.unibw.de/workshop}. \Cref{fig_collage} shows a collage of photographs from the workshop.

Based on the workshop discussions, we developed a roadmap. According to Merriam-Webster, a \textit{roadmap} is defined as \textit{a detailed plan to guide progress toward a goal}. The goal of our roadmap is to address the current challenges in small-scale testbeds for \acp{cav} and robot swarms. Our roadmap, comprising three parts, is structurally inspired by \cite{annaswamy2024control}. Part A (\cref{sec:partA}) focuses on enhancing accessibility and diversity by encouraging broad participation, promoting remote access, and ensuring inclusive data collection and design. Part B (\cref{sec:partB}) presents best practices for testbed development and maintenance, including collaboration, open-source contributions, and modular design. Part C (\cref{sec:partC}) advocates for the creation of an abstraction layer that connects testbeds to boost collaboration.

The remainder of the paper is organized as follows. We outline the current challenges in small-scale testbeds in \cref{sec:challenges}, detail our roadmap in \cref{sec:roadmap}, and draw conclusions in \cref{sec:conclusions}.

\section{Challenges} \label{sec:challenges}
The goal of our roadmap is to address the challenges in small-scale testbeds for \acp{cav} and robot swarms. We classify these challenges into two main categories. The first, \textbf{transition challenges}, refers to challenges that arise when moving from controlled, small-scale, or simulated environments to full-scale deployments. The second, \textbf{inherent real-world challenges}, comprises challenges in small-scale testbeds that are inherent in real-world operating environments, regardless of testbed fidelity---such as handling environmental unpredictability and the demands of robust distributed computing.

\subsection{Transition Challenges} \label{sec:challenges:transition}
We identified two transition challenges---\ac{sim2real} gap and small- to full-scale transition challenges---and will present them in \cref{sec:challenges:sim2real} and \cref{sec:challenges:scale}, respectively.

\subsubsection{\Ac{sim2real} Gap} \label{sec:challenges:sim2real}
\Ac{sim2real} gap hinders the applicability of technologies developed in simulation to reality. Bridging this gap is essential because it allows methods developed in simulation to be transferred reliably to the real world, ensuring that the performance observed in simulation is maintained. We identified the following aspects to bridge this gap.

\paragraph{Simulation Fidelity} Simulation fidelity refers to the degree to which a simulation replicates the behavior and characteristics of a real-world system. Digital twins improve this fidelity by integrating high-fidelity sensor models with real-time data feedback loops, thereby capturing the complex interactions between physical and virtual elements. The Mixed Cloud Control Testbed (MCCT) \cite{dong2023mixed} demonstrates this approach by merging mixed reality environments with bidirectional physical-digital synchronization, which enables detailed environmental rendering and precise replication of sensor-actuator dynamics.

\paragraph{Dynamic Environment Complexity} Real-world environments exhibit non-stationary conditions including weather variations, lighting changes, and transient obstacles. Current simulations struggle to model them, particularly their compound effects on system dynamics.

\paragraph{Multi-Agent Interaction Fidelity} Capturing emergent behaviors in \acp{cav} and robot swarms requires modeling their complex interactions. Existing approaches often oversimplify collision avoidance, failing to replicate real-world interactions.

Key points in bridging \ac{sim2real} gap include:
\begin{itemize}
    \item Ensuring sensor and actuator in simulations maintain fidelity across diverse environmental conditions and hardware configurations.
    \item Advancing domain adaptation techniques using generative approaches such as \cite{ulhas2024ganbased} to minimize perceptual discrepancies.
    \item Modeling complex interactions between \acp{cav} and robot swarms.
\end{itemize}

\subsubsection{Small-Scale to Full-Scale Transition Challenges} \label{sec:challenges:scale}
Transitioning from small- to full-scale introduces challenges in four main domains: localization, perception, actuation, and uncertainty modeling.

\paragraph{Localization} Localization in small-scale testbeds often relies on indoor positioning systems like \cite{kloock2020visionbased}, which are unsuitable for outdoor environments. Due to their limited size, small-scale testbeds cannot adopt some localization methods used in full-scale environments. For instance, radio-based systems like GPS, which are commonly used in large-scale settings, can suffer interference from onboard equipment that is placed in proximity in small-scale setups. Study \cite{schafer2023investigating} proposes pressure surface layers as a solution but necessitates further exploration for diverse settings.

\paragraph{Perception} Perception systems must overcome challenges such as different lighting conditions, blurred or obscured markers, and obstructions in the real world. Recent studies highlight the use of neural network-based perception systems trained with augmented datasets to address these variations. However, real-world robustness remains a challenge, such as out-of-distribution objects not encountered in controlled training environments.

\paragraph{Actuation} Actuation differs between small- and full-scale vehicles and robots due to varying dynamics. A proposed solution involves a translation layer that adjusts control inputs to account for the distinct dynamics of small-scale vehicles, enabling more accurate modeling of full-scale counterparts.

\begin{figure}[t]
    \centering
    \includegraphics[width=1\linewidth]{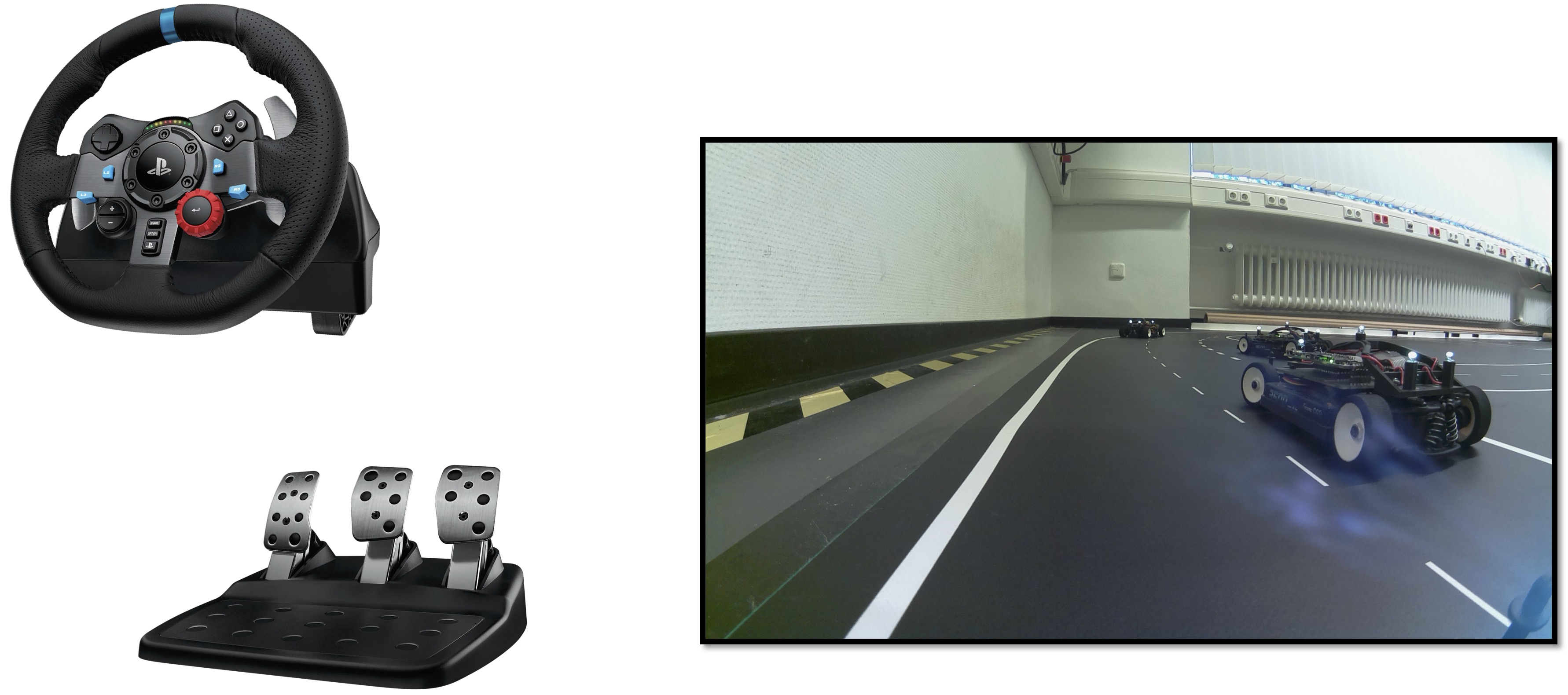}
    \caption{A case study of modeling a human-driven vehicle \cite{scheffe2023scaled}.}
    \label{fig_mixed_traffic}
\end{figure}

\paragraph{Uncertainty Modeling} Modeling uncertainties in the real world into small-scale testbeds remains challenging. These uncertainties include:
\begin{itemize}
    \item Communication Uncertainties: Communication Uncertainties refer to the unpredictable variations in data exchange between agents that arise from factors such as wireless interference, latency, and packet loss. These uncertainties often lead to a mismatch between small-scale testbeds with controlled environments and full-scale real-world scenarios, where network conditions are more complex and less predictable. In small-scale testbeds, the communication may not fully capture the disturbances in the real world. For example, network congestion can introduce delays that directly impact the performance of control algorithms for \acp{cav} and robot swarms.
    \item Mixed traffic: Small-scale testbeds often struggle to model mixed traffic. Modeling mixed traffic requires incorporating diverse traffic participants, particularly in unorganized and non-lane-based settings common in developing countries. Study \cite{scheffe2023scaled} from the \ac{cpmlab} demonstrates how human-driven vehicles can be modeled using steering controls and first-person camera feeds, as shown in \cref{fig_mixed_traffic}. Similarly, the CHARTOPOLIS testbed enhances autonomous driving tasks by incorporating human-remote-operated vehicles \cite{ulhas2022chartopolis}. Study \cite{mahbub2022platoon} proposes a framework that investigates the interactions between \acp{cav} and human-driven vehicles within a platoon. Furthermore, modeling unorganized traffic in developing regions presents additional challenges, requiring testbeds to simulate unexpected behaviors from traffic participants and diverse vehicle types.
    \item Environmental Stochasticity and System Degradation: Road friction coefficients, road surface, and vehicle payload vary in full-scale deployments. Moreover, tire wear and component aging effects also manifest differently in full-scale deployments.
\end{itemize}

Key points in small- to full-scale transition include:
\begin{itemize}
    \item Designing localization and perception systems that maintain functional consistency across different scales and diverse environmental conditions.
    \item Developing scale-agnostic control interfaces that compensate for dynamic mismatches.
    \item Creating validation metrics for cross-scale transferability of uncertainty models.
    \item Modeling uncertainties such as communication uncertainties, mixed traffic, environmental stochasticity, and system degradation in small-scale testbeds.
\end{itemize}

\subsection{Inherent Real-World Challenges} \label{sec:challenges:real-world}
Inherent real-world challenges persist in small-scale testbeds due to fundamental characteristics of physical environments. We focus on two critical challenges: environmental unpredictability and demands of robust distributed computing, both requiring fundamental advances in testbed design and algorithmic development.

\subsubsection{Environmental Unpredictability} \label{sec:challenges:environment}
Real-world environments exhibit three forms of uncertainty: mixed traffic, anomalous object interactions, and contextual variability.

\paragraph{Mixed Traffic}
Real-world environments are inherently unpredictable due to coexistence with human-operated vehicles, which introduce stochastic and heterogeneous decision-making behaviors. Autonomous systems operating in such settings must perform real-time behavior inference and risk-aware motion planning under incomplete and noisy observations. These challenges become substantially more severe in highly unstructured traffic scenarios common in developing regions, including dense urban areas or rural areas in countries such as Egypt, India, and China. In these environments, traffic flow often lacks strict lane discipline and is partially governed by local social conventions, right-of-way rules are loosely enforced or implicitly negotiated, and road users include heterogeneous agents such as motorcycles, auto-rickshaws, bicycles, pedestrians, animal-drawn carts, and informal public transport vehicles. These characteristics are generally not considered by the autonomous systems developed in developed countries. In these regions, intention inference becomes ambiguous, and trajectory prediction errors grow significantly, thereby increasing collision risk and reducing algorithmic reliability.

\paragraph{Anomalous Object Interactions}
Current perception systems remain vulnerable to out-of-distribution obstacles ranging from temporary construction zones to unconventional vehicle types and irregularly shaped loads. Such objects are especially common in developing regions, where vehicle modifications, informal trailers, and roadside commerce introduce perception distributions absent from most benchmark datasets. Therefore, robust operation requires generalization beyond training data obtained from controlled small-scale testbeds or real-world environments with limited diversity.

\paragraph{Environmental Variability}
Transient environmental factors, such as sudden weather changes, temporary obstructions, and ad-hoc traffic redirection by human controllers, create non-stationary operating contexts that challenge both sensing and control systems.

Key points in environmental unpredictability include:
\begin{itemize}
    \item Developing multi-agent behavior models that capture unstructured interactions, especially in non-lane-based traffic scenarios.
    \item Creating perception systems robust to unusual object categories and extreme occlusion scenarios.
    \item Handling non-stationary environmental conditions common in the real world.
\end{itemize}

\subsubsection{Distributed Computing Demands} \label{sec:challenges:distributed}
Distributed computing in \acp{cav} and robot swarms is characterized by three primary challenges.

\paragraph{Resource Constraints, Real-Time Computing, and Fault Tolerance}
The operational scenarios often demand real-time processing to ensure rapid decision-making. However, on-board computing units in \acp{cav} and individual robots are usually limited in processing power, memory, and energy, rendering real-time computing in these distributed systems challenging. Moreover, these distributed systems require robust fault tolerance mechanisms to handle node failures and communication disruptions to ensure system safety and performance.

\paragraph{Communication Overhead and Synchronization}
Distributed tasks require frequent data exchanges to achieve coordination. However, wireless communication is subject to latency, limited bandwidth, and potential interference, which hinder timely synchronization and thus degrade the performance of distributed control algorithms.

Key points in distributed computing demands include:
\begin{itemize}
    \item Developing distributed algorithms that are efficient under strict resource limitations.
    \item Integrating fault tolerance and real-time processing capabilities to maintain system performance in the presence of disruptions.
    \item Managing communication overhead while ensuring reliable and timely synchronization across the network.
\end{itemize}

\section{Roadmap} \label{sec:roadmap}
In this section, we delineate our roadmap to address the challenges in small-scale testbeds for \acp{cav} and swarm robots, as proposed in \cref{sec:challenges}. Note that our roadmap aims to synthesize community-identified research needs and future directions, rather than proposing new methodological or algorithmic contributions. Our roadmap contains three parts. We propose the first part in \cref{sec:partA}, which underscores the importance of enhancing the accessibility and diversity of testbeds. The second part, described in \cref{sec:partB}, shares best practices for testbeds, covering a broad spectrum, including collaboration, knowledge sharing, open-sourcing, modularity, etc. In the third part, presented in \cref{sec:partC}, we suggest developing an abstraction layer to interconnect testbeds, which can enable collaboration, maintain consistency, and drive innovation.

\subsection{Part A: Enhancing Accessibility and Diversity}\label{sec:partA}
We highlight the important role that testbeds play in fostering accessibility and diversity within \ac{its} and robot swarms, particularly for underrepresented and geographically distributed communities. Enhanced accessibility encourages more research groups to participate, promoting a collective approach to the current challenges. This variety helps reduce the gap between controlled testbed environments and real-world deployments.

In \cref{sec:accessibility:research}, we describe the role of testbeds in research and education. \Cref{sec:engagement} shares ways to encourage engagement. \Cref{sec:accessibility:remote} details how remote access expands participation. In \cref{sec:accessibility:diversity}, we examine methods to enhance diversity in \ac{its} research, and \cref{sec:accessibility:small-scale} explores the use of small-scale vehicles to engage local expertise and validate advanced technologies.

\subsubsection{Testbeds for Research and Education} \label{sec:accessibility:research}
\begin{figure}
    \centering
    \begin{subfigure}[t]{1\linewidth}
        \centering
        \includegraphics[width=\linewidth]{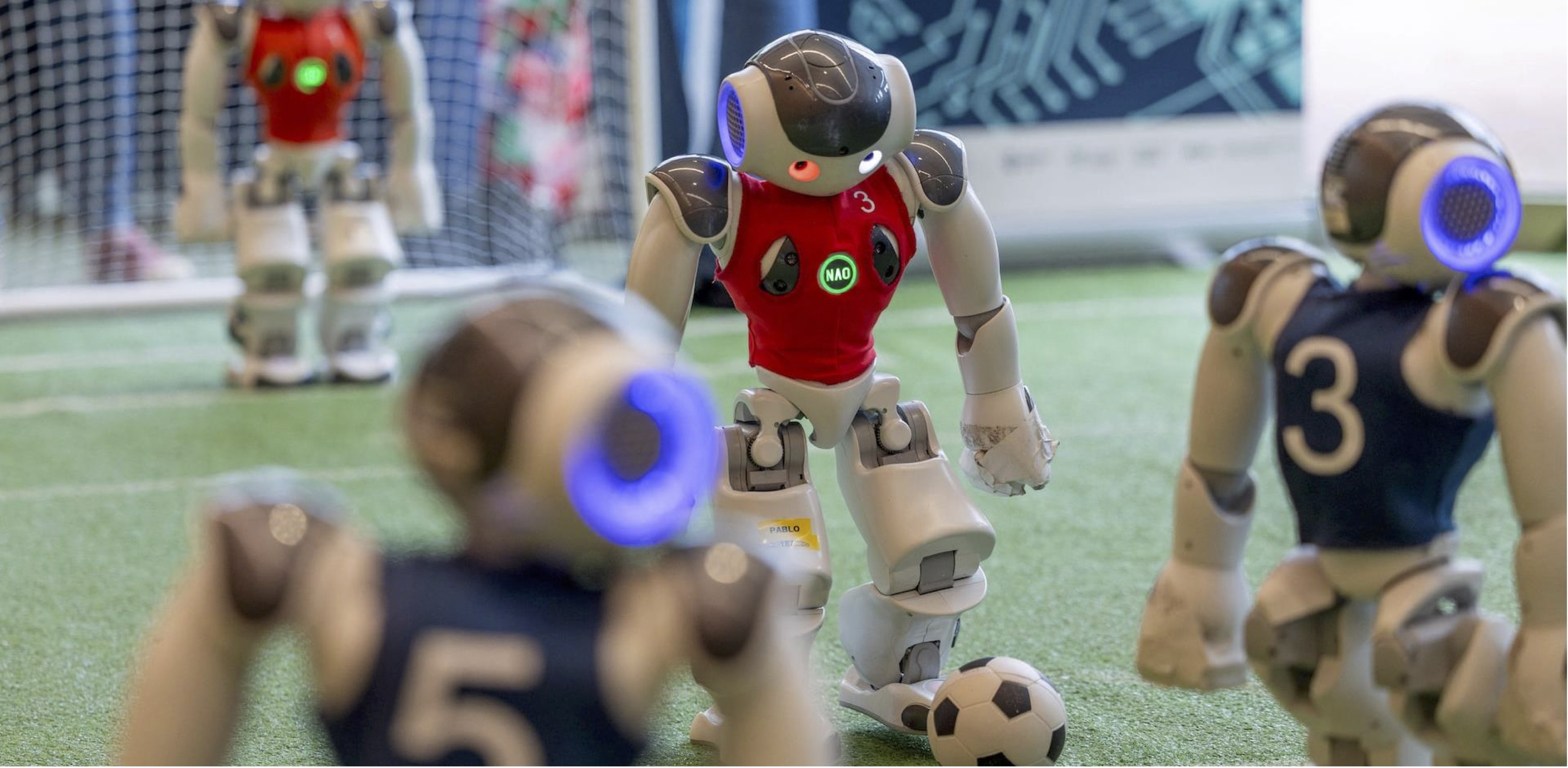}
        \caption{RoboCup at ETH Zürich.}\label{fig_RoboCup}
    \end{subfigure}

    \medskip
    
    \begin{subfigure}[t]{1\linewidth}
        \centering
        \includegraphics[width=\linewidth]{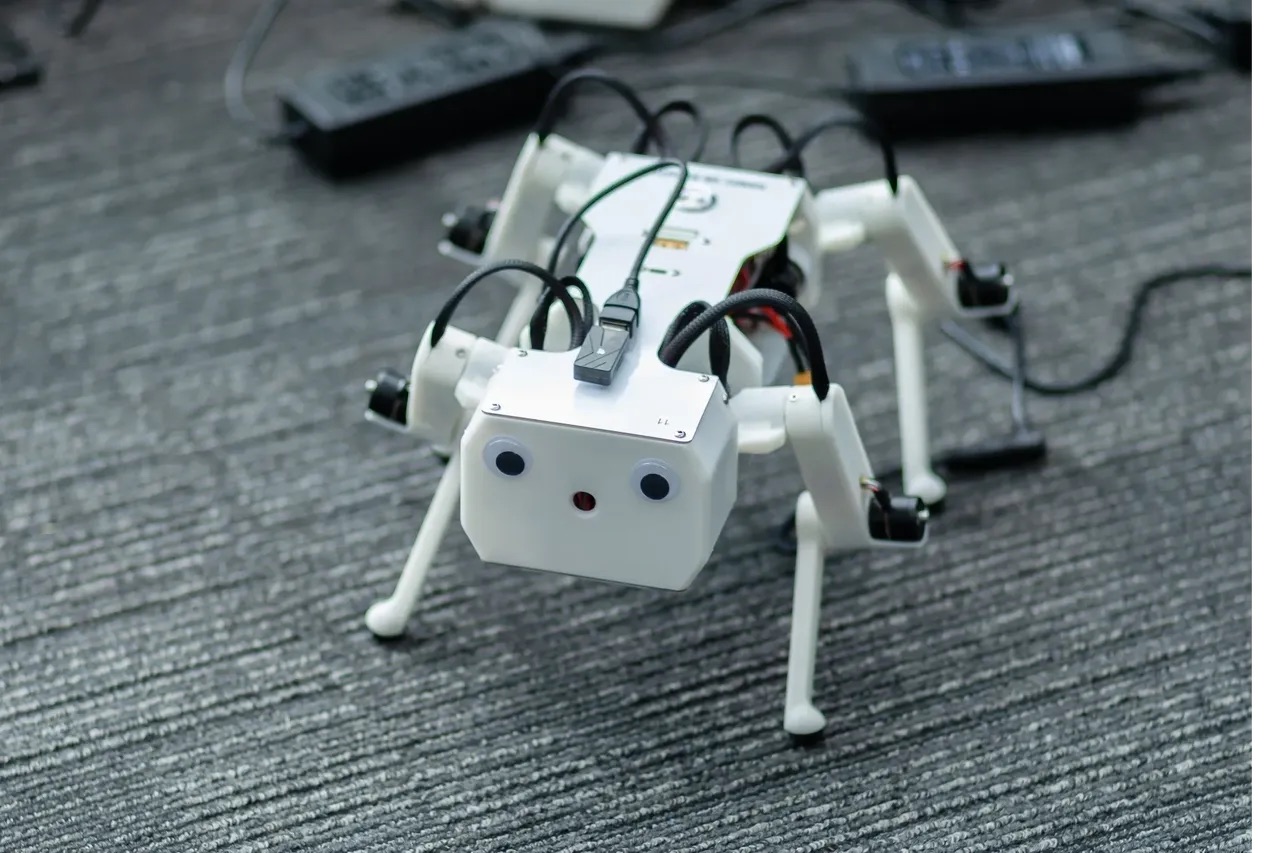}
        \caption{Hands-On Robotics, which builds on the Stanford Pupper---a small-scale and low-cost quadruped robot.} \label{fig_handsonrobot}
    \end{subfigure}
    \caption{
        Case studies of low-cost education platforms.
    }\label{fig_low_cost_platforms}
\end{figure}

Testbeds are essential for advancing research and education by providing platforms for practical experiments on real systems. By providing hands-on opportunities, testbeds contribute to educating the next generation of engineers and promote inclusivity in both research and education. We identified two key measures that can enhance testbeds' accessibility.
The first measure is cost-effectiveness. Low-cost platforms, such as \textit{Hands-On Robotics}\footnote{\href{https://handsonrobotics.org}{https://handsonrobotics.org}} and \textit{RoboCup}\footnote{\href{https://nomadz.ethz.ch}{https://nomadz.ethz.ch}} shown in \cref{fig_low_cost_platforms}, leverage small-scale robots. These platforms are accessible for both research and education, enabling engagement with advanced technologies at a low  cost. The second measure is lightweight implementation. Educational testbeds should emphasize simplicity and resource efficiency. Designing testbeds with simplicity allows students with limited knowledge and experience to engage with research and education activities. This idea is similar to the University of Pennsylvania's \textit{TinyML} course\footnote{\href{https://tinyml.seas.upenn.edu}{https://tinyml.seas.upenn.edu}}, which exemplifies how resource-efficient platforms can teach fundamental principles of machine learning, enabling widespread accessibility for students. These small-scale platforms provide an important advantage that cannot be fully achieved by simulation-only tools such as CARLA \cite{dosovitskiy2017carla}, VISSIM \cite{fellendorf2010microscopic}, or CarSim \cite{benekohal1988carsim}. While these simulators offer high-fidelity environment and traffic modeling and support configurable models of sensing, actuation, and timing imperfections, they rely on parameterized models that cannot fully capture emergent effects arising from real hardware and software stacks. In contrast, small-scale testbeds enable students and researchers to develop and evaluate algorithms on physical systems. Since algorithms ultimately operate on real hardware, these platforms expose users to sensing imperfections, actuator limitations, real-time constraints, and system-integration effects that naturally arise in practice. As a result, small-scale testbeds serve as an essential intermediate validation layer between simulation and full-scale deployment, supporting more complete education and helping to narrow the gap between theoretical development and practical implementation. The two key measures result in the following key points:
\begin{itemize}
    \item Balancing the trade-off between the complexity required for advanced research and the simplicity necessary for ensuring accessibility and usability for students.
    \item Developing approaches for leveraging testbeds to facilitate collaboration between academia and industry, thereby creating opportunities for students to engage with industry professionals.
\end{itemize}

\subsubsection{Encouraging Engagement}\label{sec:engagement}

Fostering engagement among researchers, educators, and students in developing small-scale testbeds is essential for advancing \acp{cav} and robot swarms. In this section, we share strategies to achieve this, e.g., educational programs, competitions, workshops and seminars, and international collaborations.

\paragraph{Educational Programs}

Incorporating small-scale testbeds into educational programs equips students with hands-on skills to tackle real-world challenges. Case studies include:
\begin{itemize}
    \item RoboRacer \cite{o2020f1tenth}: RoboRacer\footnote{\href{https://roboracer.ai}{https://roboracer.ai}}, also known as F1TENTH, offers autonomous racing courses across multiple universities. These courses engage students in developing control, perception, and planning algorithms tailored for autonomous vehicles.
    \item \ac{cpmlab} \cite{kloock2021cyberphysical, kloock2023architecture, scheffe2020networked, kloock2023testing, mokhtarian2023cpm, mokhtarian2022remote, mokhtarian2022cpma, mokhtarian2022cpm}: The \ac{cpmlab} offers a multi-level educational framework consisting of
    \begin{enumerate*}
        \item \textit{CPM Academy}\footnote{\href{https://cpm-remote.lrt.unibw-muenchen.de/academy}{https://cpm-remote.lrt.unibw-muenchen.de/academy}} \cite{mokhtarian2022remote, mokhtarian2022cpma, mokhtarian2022cpm} for undergraduates,
        \item a graduate's course\footnote{\href{https://cpm.lrt.unibw.de/course-materials}{https://cpm.lrt.unibw.de/course-materials}}, and
        \item a PhD course\footnote{\href{https://cpm.lrt.unibw.de/EECI-IGSC2021}{https://cpm.lrt.unibw.de/EECI-IGSC2021}}.
    \end{enumerate*}
    These courses prepare students for the challenges in \acp{cav}.
\end{itemize}

These small-scale testbeds combine theoretical knowledge with practical applications, preparing young generation to contribute effectively to advancements in \ac{its}.

\paragraph{Testbeds for Competitions}

Competitions incentivize innovations. By participating in such events, students and researchers can refine their technical skills and collaborate within a competitive atmosphere. Identified initiatives include: 
\begin{itemize}
    \item RoboRacer \cite{o2020f1tenth}: This autonomous racing competition challenges participants to create high-performance algorithms for fast and precise vehicle control.
    \item ForzaETH Race Stack \cite{baumann2024forzaeth}: A modular software platform\footnote{\href{https://github.com/ForzaETH/race\_stack}{https://github.com/ForzaETH/race\_stack}} tailored for RoboRacer competitions, enabling customization and adaptability to diverse environmental conditions.
    \item CPM Olympics \cite{mokhtarian2022cpm}: \ac{cav} motion-planning competitions organized by the \ac{cpmlab}.
\end{itemize}

These competitions foster creativity, teamwork, and technical expertise, while advancing \ac{its} technologies.

\paragraph{Workshops and Seminars}

Workshops and seminars foster collaboration and knowledge exchange. They provide a forum for established experts and early-career researchers to share insights, discuss emerging trends, and address challenges in small-scale testbeds. Besides our workshop, there is another recent workshop held on September 24 at the IEEE Intelligent Transportation Systems Conference (ITSC) 2024, called \textit{Small-Scale Testbeds for Connected and Automated Vehicles in Research and Education}\footnote{\href{https://kth-sml.github.io/itsc24-workshops/\#small-scale}{https://kth-sml.github.io/itsc24-workshops/\#small-scale}}.

\paragraph{International Research Initiatives}
Collaborative research across universities and institutions from different countries contributes to addressing the diverse challenges in \acp{cav} and robot swarms. Countries such as India, South Africa, and Egypt face common issues with diverse, non-lane-based traffic scenarios, making them ideal collaborators for the development
and evaluation of intelligent transportation solutions.

One initiative is \textit{CPM Global}\footnote{\href{https://www.unibw.de/cas-en/news/cpm-global-workshop}{https://www.unibw.de/cas-en/news/cpm-global-workshop}}, a workshop initiated by Prof. Bassam Alrifaee. This collaboration unites researchers from Germany, South Africa, Egypt, and India to develop small-scale testbeds tailored to address regional challenges in \ac{its}. \textit{CPM Global} aims to establish a unified framework for testbed research, promoting inclusivity and innovation in solving global \ac{its} problems.

Another initiative is the workshop \textit{Advancing Connected Autonomous Vehicle Technologies for the Indian Ecosystem}, held on December 10, 2024, at the Indian Institute of Technology Bombay. It represented a focused effort to address region-specific challenges in \ac{its}.
By aligning academic research with practical considerations unique to India’s transportation environment, the workshop underscores the significance of locally tailored solutions in advancing \ac{its}.

\subsubsection{Remote Access} \label{sec:accessibility:remote}
\begin{figure}[t]
    \centering
    \includegraphics[width=1\linewidth]{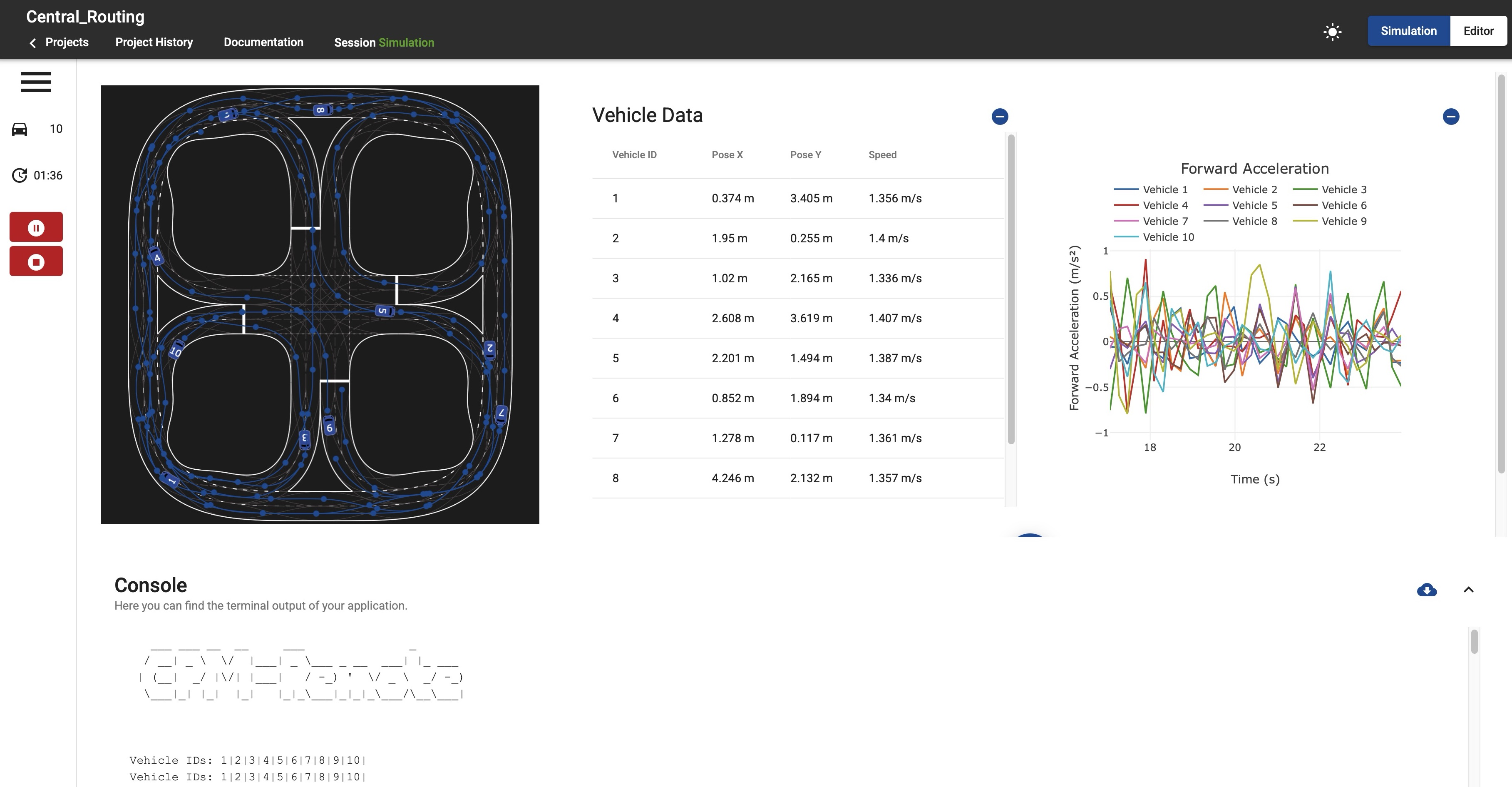}
    \caption{A case study of remote access (CPM Remote \cite{mokhtarian2022remote}) of a testbed (the \ac{cpmlab} \cite{kloock2021cyberphysical}).}
    \label{fig_cpm_remote}
\end{figure}

Remote access to testbeds increases their accessibility, particularly for underrepresented communities and those in geographically remote areas. Besides, providing cost-effective remote platforms allows researchers to engage without physical presence. An example of this is the \ac{cpmlab}\footnote{\href{https://cpm.lrt.unibw.de}{https://cpm.lrt.unibw.de}} \cite{kloock2021cyberphysical}, which provides remote access to its testbed via \textit{CPM Remote}\footnote{\href{https://cpm-remote.lrt.unibw-muenchen.de}{https://cpm-remote.lrt.unibw-muenchen.de}} \cite{mokhtarian2022remote}, as shown in \cref{fig_cpm_remote}. Another example is Duckietown \cite{paull2017duckietown}, a small-scale testbed for robot swarms, which offers remote access through ZeroTier. Such initiatives foster global collaboration and inclusion. Key points regarding remote access include:
\begin{itemize}
    \item Enhancing remote access systems to better support users with limited technical expertise.
    \item Developing and implementing measures to guarantee equitable access to remote testbed platforms.
\end{itemize}

\subsubsection{Diversity in ITS Research} \label{sec:accessibility:diversity}
Transportation systems vary significantly based on factors such as geography, culture, and infrastructure. Addressing this diversity is essential for validating \ac{its} technologies across different environments in various countries. We examine the multifaceted differences of diversity in \ac{its} research by identifying barriers that hinder the application of advanced technologies in real-world settings and presenting relevant case studies of initiatives.

\paragraph{Identifying Barriers} A major challenge is the gap between advanced \ac{its} technologies and real-world environments, especially in developing regions. Local researchers are essential in recognizing region-specific challenges and enhancing data collection, thereby providing valuable insights for tailored solutions. Bridging the gap between state-of-the-art research and practical transportation needs can promote effective international collaboration and ensure that technological advances are applicable across diverse regions.
    
\paragraph{Dataset Diversity} Dataset diversity is essential for \ac{its}, yet many current datasets are predominantly collected in developed countries and reflect limited environmental and traffic conditions. This imbalance limits the applicability of \ac{its} technologies in diverse settings. Expanding data collection to include developing countries ensures a more accurate global representation and enables the development of universally applicable technologies. Encouraging more engagement from local researchers is essential to address this imbalance and can significantly contribute to addressing the challenges in \cref{sec:challenges:environment}. 

\paragraph{Tailored Testbed Design} In addition to improving data collection, another promising solution is the development of tailored small-scale testbeds that replicate local traffic scenarios typical of developing regions. These testbeds, possibly along with remote access for resource‐constrained communities, can support inclusive research within the community.
Key design principles include:
\begin{itemize}
    \item Identifying challenges that researchers in developing countries face during data collection, including issues related to regulations and legal requirements.
    \item Enabling adaptability of small-scale testbeds to reflect changes in the road infrastructure in developing countries.
    \item Developing testbeds that can accurately replicate environmental factors such as road quality and traffic density in developing countries
\end{itemize}

\paragraph{Case Studies} The \textit{IEEE ITSS Young Professionals (YP)} committee\footnote{\href{https://ieee-itss.org/yp}{https://ieee-itss.org/yp}} supports early-career engineers, scientists, and technical experts by providing opportunities for networking, collaboration, and professional development within the \ac{its} community, helping members launch their careers and build connections across academia and industry. The \textit{IEEE ITSS Women and Underrepresented Groups (Wi-ITS)} committee\footnote{\href{https://ieee-itss.org/wie}{https://ieee-itss.org/wie}}, established in 2023, promotes the visibility, participation, and professional advancement of women and other underrepresented groups in \ac{its} by developing programs and activities that support career growth and leadership development across the society.

\subsubsection{Small-Scale Vehicles: A Technology Enabler for Developing Countries} \label{sec:accessibility:small-scale} Small-scale vehicles, available in various sizes, address specific research challenges and enhance \ac{cav} applicability. They are especially important for resource-constrained regions, allowing local researchers to test globally developed algorithms in local contexts and build regional technical expertise. The primary scales include:
\begin{itemize}
    \item 1:18 to 1:16 (Hand-Held): These vehicles, equipped with onboard processors, validate cooperative vehicle algorithms, often using overhead vision tracking systems, as seen in MRS Scaled City \cite{mahdi2024mrs}.
    \item 1:4: Equipped with processors, sensors, and actuators, vehicles with this scale support complex systems and automotive-standard communication protocols like CAN bus, bridging small-scale testbeds and real-world applications \cite{reiter2014model}.
    \item 1:2: The Electric Vehicles Rally (EVER\footnote{\href{https://www.linkedin.com/company/ever-electric-vehicle-rally}{https://www.linkedin.com/company/ever-electric-vehicle-rally}}) designs one-passenger vehicles, fostering innovation and skill development. Testing advanced driver-assistance systems at this scale refines algorithms and builds participant expertise under varied conditions.
\end{itemize}

Key points related to small-scale vehicles include:
\begin{itemize}
    \item Identifying design changes and technological adjustments needed for small-scale vehicles to operate on unpaved or uneven roads and with limited infrastructure in developing regions.
    \item Engaging local researchers in developing countries.
    \item Exploring ways in which small-scale vehicles can act as platforms for adapting and evaluating globally developed algorithms in local conditions.
\end{itemize}

\subsection{Part B: Sharing Best Practices}\label{sec:partB}
To address the current challenges in small-scale testbeds for \acp{cav} and robot swarms, we need best practices to ensure effectiveness, scalability, and usability when developing and maintaining them. This section presents these best practices. In \cref{sec:best-practices:sharing}, we discuss the importance of collaboration and knowledge sharing among researchers. \Cref{sec:best-practices:open-source} highlights mechanisms to promote open-source contributions. In \cref{sec:best-practices:modular-testbeds}, we suggest the design of purpose-driven and modular testbeds. \Cref{sec:best-practices:usability} covers approaches to streamline usability and development, while \cref{sec:best-practices:interoperability} focuses on improving interoperability and communication. Finally, \cref{sec:best-practices:continuous-development} emphasizes the need for continuous development to maintain the relevance of testbeds.

\subsubsection{Fostering Collaboration and Knowledge Sharing} \label{sec:best-practices:sharing}
Collaboration among researchers enhances testbed utility. A centralized hub for sharing code, tools, and documentation fosters this collaboration. Detailed documentation helps users understand a testbed's capabilities and limitations, guiding customization for specific needs. Customizable testbeds enhance flexibility, enabling space for innovation by other researchers. Moreover, standardized benchmarking scenarios facilitate comparisons of algorithms and accelerate progress in the field.

\subsubsection{Encouraging Open-Source Contributions} \label{sec:best-practices:open-source}
Open-source frameworks and tools are instrumental in promoting accessibility and collaboration. However, their development often requires additional efforts to ensure that the resulting code is reusable, adaptable, and maintainable. We discovered two mechanisms to incentivize open-source contributions:
\begin{enumerate*}
    \item Pull Mechanism: incentivizing open-source development through competition tracks or academic recognition, and
    \item Push Mechanism: requiring open-source releases as a condition for eligibility in competitions or funding programs, ensuring broad community benefits.
\end{enumerate*}

\subsubsection{Designing Purpose-Driven and Modular Testbeds} \label{sec:best-practices:modular-testbeds}
Given the resource-intensive nature of testbed development, their design must align with research objectives while remaining flexible for updates. Purpose-driven design ensures relevance in addressing emerging challenges in \ac{cav} and robot swarms. In addition, modular hardware and software, supported by platforms like Arduino, Raspberry Pi, and \ac{ros}, enable the integration of diverse components like sensors and communication modules. This fosters flexibility for evolving research needs.

\subsubsection{Streamlining Usability and Development} \label{sec:best-practices:usability}

The usability of testbeds can be enhanced through pre-configured, ready-to-use modules that allow researchers and students to focus on experimentation rather than setup. These ``batteries-included'' solutions reduce development time and provide baseline implementations for benchmarking. Moreover, containerized software environments can further simplify testbed deployment and operation. By ensuring reproducibility and lowering the entry barrier, containerization facilitates consistent experimentation across diverse research groups. Additionally, regular evaluations and updates to testbeds are necessary to address obsolescence, particularly in low-cost platforms that may lag behind advancements in computational or sensing capabilities.

\subsubsection{Improving Interoperability and Communication} \label{sec:best-practices:interoperability}

Creating interoperability between testbeds enables collaborative research and broader applicability. Designing well-defined communication interfaces and selecting appropriate middleware can minimize technical debt and ensure compatibility with future developments. These considerations are particularly important for integrating testbeds into industrial and academic ecosystems, where seamless data exchange and collaboration are essential.

\subsubsection{Continuous Development} \label{sec:best-practices:continuous-development}
Testbeds need to be evaluated regularly for their suitability to allow benchmarking of current research challenges. Low-cost testbeds, in particular, are susceptible to becoming outdated in specific areas. For instance, the limited computational power available in the RoboCup Standard Platform League (SPL)\footnote{\href{https://spl.robocup.org}{https://spl.robocup.org}} currently constrains progress in integrating state-of-the-art computer vision techniques. Therefore, continuous upgrades are necessary to keep testbeds aligned with evolving research challenges.

Key points related to best practices include:
\begin{itemize}
    \item Fostering collaboration, knowledge sharing, and standardization to improve testbed effectiveness.
    \item Promoting open-source contributions through competitive incentives and funding requirements.
    \item Designing testbeds to be purpose-driven, modular, and adaptable to evolving research needs.
    \item Simplifying usability and development through pre-configured modules and containerized environments.
    \item Improving interoperability and communication for seamless integration across research and industrial systems.
    \item Ensuring continuous development and frequent updates to maintain testbed relevance.
\end{itemize}

\subsection{Part C: Global Testbed Connectivity} \label{sec:partC}
Global testbed connectivity aims to connect testbeds across regions. Achieving this connectivity supports collaboration at scale, increases reproducibility, and enables the study of complex scenarios that no single testbed can represent alone.

\subsubsection{Abstraction Layers}
\begin{figure}[t]
    \centering
    \includegraphics[width=1\linewidth]{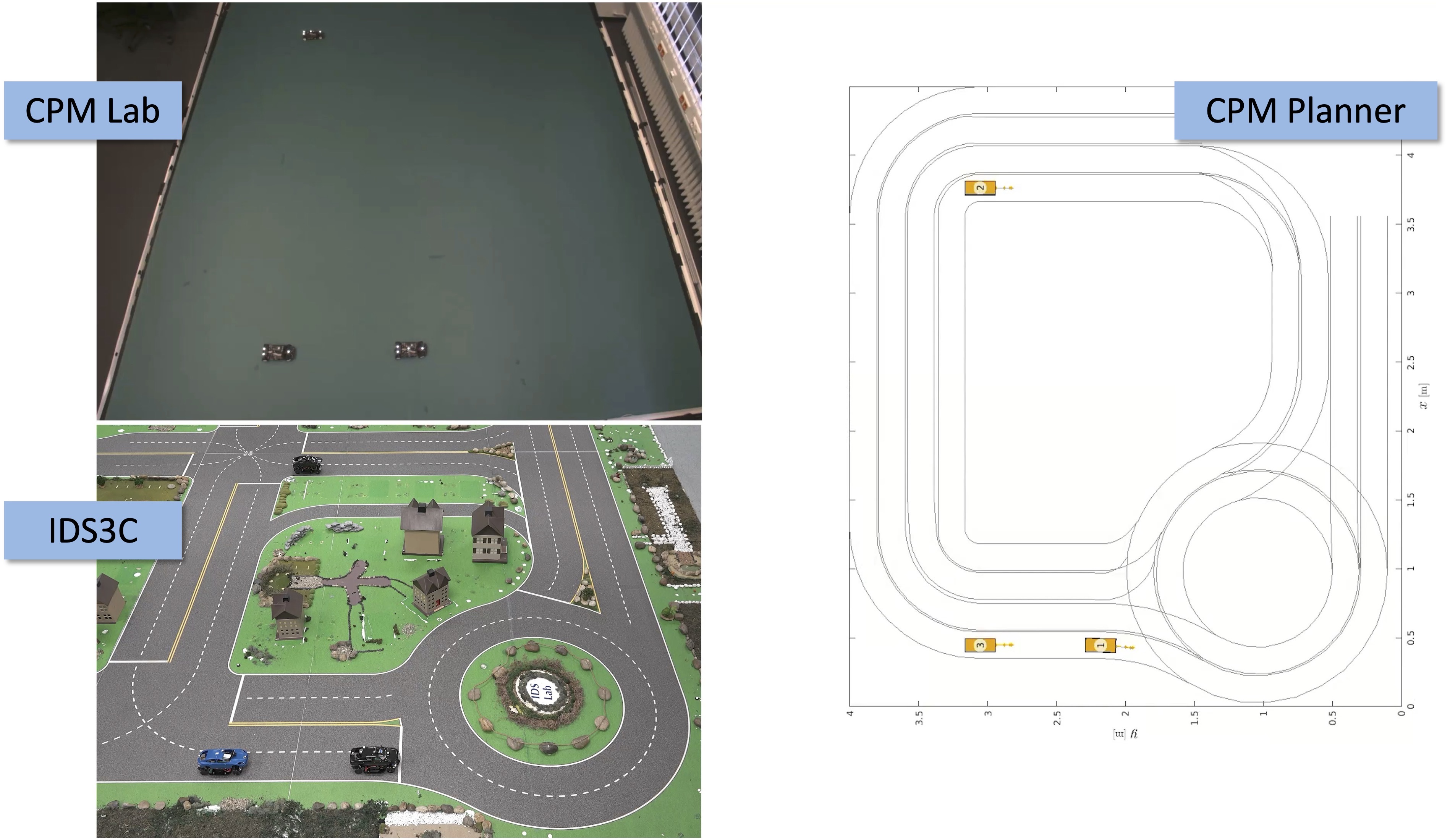}
    \caption{A case study demonstrating an abstraction layer connecting the \ac{cpmlab} \cite{kloock2021cyberphysical} and IDS\textsuperscript{3}C \cite{chalaki2022research}.}
    \label{fig_connected_tetbeds}
\end{figure}

We identified establishing an abstraction layer to be one promising approach. An abstraction layer that connects diverse testbeds can greatly enable global testbed connectivity, maintain consistency, and drive innovation. This abstraction layer acts as a mediator that allows distinct testbeds---each with unique setups---to communicate and interact effectively. The abstraction layer not only simplifies collaboration but also accelerates innovation by allowing testbeds to complement each other. Researchers can explore interactions between diverse systems, validate algorithms across varied setups, and address complex challenges that would be difficult to replicate within a single testbed. By adopting such an abstraction layer, the research community can unlock new opportunities for collaboration and innovation, advancing the field of \acp{cav} and robot swarms through shared resources and collective expertise. A practical \textbf{case study} of this concept is connecting the \ac{cpmlab} \cite{kloock2021cyberphysical} and the Information and Decision Science Lab Scaled Smart City (IDS\textsuperscript{3}C) \cite{chalaki2022research}. Each testbed is unique, e.g., in scale, size, and \ac{cav} setup, but they have in common a communication of motion planner and \acp{cav} via WiFi. Both testbeds were connected through our proposed abstraction layer using a service-based interface to read measurements and apply motion plans. The same motion planner \cite{scheffe2022receding} was deployed unchanged on both testbeds and executed in identical multi-vehicle intersection scenarios with three \acp{cav}. Via the abstraction layer, the motion planner measures the vehicle states at each time step and transmits new motion plans back to the respective testbed controllers for execution. Runtime measurements collected on both testbeds indicate that the additional overhead introduced by the abstraction layer remains within a few milliseconds, with message-level delays in the sub-millisecond range, and total time per control cycle below $\SI{40}{\milli\second}$. These results demonstrate that platform-independent deployment of motion planners can be achieved by an abstraction layer without compromising real-time feasibility. \Cref{fig_connected_tetbeds} illustrates the experimental setup, where the motion planner developed in the \ac{cpmlab} successfully controlled the vehicles in IDS\textsuperscript{3}C through the proposed abstraction layer.

\subsubsection{Standardized Framework}
Global testbed connectivity benefits from a standardized framework that provides compatibility and consistency across testbeds, independent of the specific testbed-connecting method used. Such a framework establishes common expectations for communication, data exchange, and coordination. Key principles include:
\begin{itemize}
    \item Unified Interfaces: Unified interfaces facilitate the connection of diverse testbeds. By employing unified interfaces, researchers can efficiently compare results, replicate experiments, and extend existing work.
    \item Integrated Connectivity: A common communication interface enables multiple testbeds to combine functionalities and simulate complex scenarios. This connection draws on the specialized expertise of different research groups to jointly develop advanced systems.
    \item Interdisciplinary Cooperation: The framework encourages collaboration across fields such as computer science, mechanical engineering, and electrical engineering, thereby enriching research with diverse perspectives.
    \item Augmented Capabilities: Interconnecting testbeds allows researchers to merge individual strengths and address platform-specific limitations. For instance, integrating a testbed that supports the modeling of human-driven vehicles with one focused on planning and control facilitates the study of mixed traffic for \acp{cav}.
\end{itemize}

Key research directions related to global testbed connectivity include:
\begin{itemize}
    \item Designing scalable testbed-connecting methods, including abstraction layers, that maintain performance and flexibility as additional testbeds are incorporated.
    \item Developing and validating standardized interfaces and communication protocols for seamless multi-testbed coordination.
    \item Exploring methods that allow testbed-connecting methods to adapt dynamically to changes in hardware, software, and research requirements.
    \item Conducting detailed case studies to evaluate testbed-connecting methods and establish practical guidelines for multi-testbed experimentation.
\end{itemize}

To consolidate our roadmap, we provide representative evaluation metrics related to testbed fidelity, cross-scale transfer, and global testbed connectivity mechanisms in \cref{tab:metrics}. These metrics include, for example, sensing and actuation latency, communication packet loss, and environmental variability for assessing testbed fidelity; behavioral consistency, safety-related outcomes, and real-time feasibility for evaluating cross-scale transfer; and interface generality, runtime overhead, scalability, and deployment portability for characterizing global testbed connectivity mechanisms. Note that the listed metrics are intended as illustrative guidance rather than a prescriptive or exhaustive benchmark.

\begin{table*}[t]
\centering
\caption{Evaluation dimensions and example metrics for testbed fidelity, cross-scale transfer, and global testbed connectivity.}
\label{tab:metrics}
\begin{tabular}{p{0.13\linewidth} p{0.16\linewidth} p{0.62\linewidth}}
\toprule
\textbf{Evaluation Aspect} & \textbf{Dimension} & \textbf{Example Metrics} \\
\midrule
\multirow{4}{*}{Testbed Fidelity}
& Sensing fidelity & Noise characteristics, latency, dropouts/occlusion rate, robustness under lighting changes \\
& Actuation fidelity & Control latency, tracking error, saturation behavior, repeatability \\
& Environment realism & Scenario variability, obstacle diversity, traffic participant heterogeneity \\
& Communication realism & Packet loss rate, latency distribution, bandwidth variability \\
\midrule
\multirow{4}{*}{\makecell[l]{Cross-Scale\\Transfer}}
& Behavioral consistency & Similarity of trajectories, collision rates, near-miss statistics (e.g., minimum inter-vehicle distance) \\
& Control performance & Tracking error, safety margin violations, real-time feasibility (e.g., deadline miss rate) \\
& Sensitivity to uncertainty & Performance degradation under sensing, actuation, communication, and environment variability \\
& Transfer effort & Required code modification, parameter retuning effort \\
\midrule
\multirow{4}{*}{\makecell[l]{Global Testbed\\Connectivity}}
& Interface generality & Coverage of perception, planning, and control interfaces across testbeds \\
& Runtime overhead & End-to-end latency, increase in control-cycle time \\
& Scalability & Number of supported agents and testbeds, communication load growth \\
& Deployment portability & Ability to deploy identical algorithms across platforms without modification \\
\bottomrule
\end{tabular}
\end{table*}

\section{Challenges and Roadmap: Traceability and Interrelations}\label{sec:challenges-roadmap}
In \cref{sec:challenges}, we presented the four identified challenges that motivate the roadmap, and in \cref{sec:roadmap}, we presented the detailed three parts of our roadmap. To make the link between the challenges and the roadmap parts explicit, we summarize the traceability between the challenges and the roadmap parts in \cref{sec:challenges-roadmap:traceability} and relations between the roadmap parts in \cref{sec:challenges-roadmap:interrelations}.

\subsection{Challenge-to-Roadmap Traceability}\label{sec:challenges-roadmap:traceability}
\begin{figure}[t]
    \centering
    \includegraphics[width=0.4\textwidth]{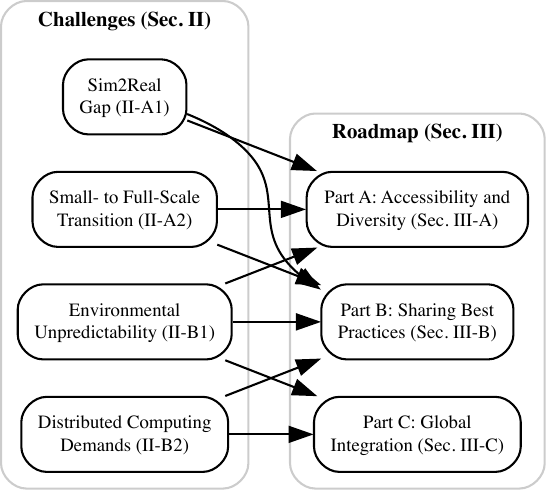}
    \caption{Traceability between the four challenges identified  in \cref{sec:challenges} and the three roadmap parts presented in \cref{sec:roadmap}. Each arrow indicates that the roadmap part contains measures that address the challenge.}
    \label{fig_challenges_roadmap_relation}
\end{figure}

\Cref{fig_challenges_roadmap_relation} maps each identified challenge in \cref{sec:challenges} to the three roadmap parts in \cref{sec:roadmap}. The figure serves as a compact traceability view: each arrow indicates that the corresponding roadmap part contains measures that directly address the challenge or reduce its impact in practice. This traceability view also clarifies that the roadmap parts are not independent. For example, roadmap Part A increases access to physical experimentation and broadens traffic scenario coverage, which supports evaluation under diverse traffic conditions and helps reduce the mismatch between controlled testbeds and the real world. Roadmap Part B improves development practice and interoperability, which supports reproducible experiments and scalable system integration when algorithms, software stacks, and hardware configurations evolve. Part C connects testbeds, which supports cross-site validation and coordinated experimentation and enables studying traffic scenarios that a single testbed cannot represent alone.

\subsection{Roadmap Interrelations}\label{sec:challenges-roadmap:interrelations}
\begin{figure}[t]
    \centering
    \includegraphics[width=0.75\linewidth]{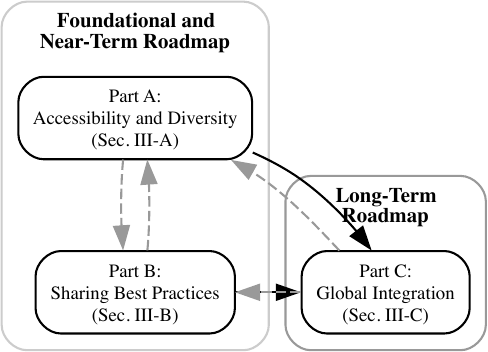}
    \caption{Part-level interrelations of the roadmap. Solid arrows indicate prerequisites for long-term global testbed connectivity. Gray dashed arrows indicate reinforcing couplings or feedback.}
    \label{fig_roadmap_parts_relation}
\end{figure}

\Cref{fig_roadmap_parts_relation} provides a part-level view that depicts the interrelations between the three parts of our roadmap. We treat Parts A and B as foundational and near-term because they lower entry barriers, improve reproducibility, and establish shared development practices that can be adopted by individual testbeds without requiring global testbed connectivity. Part A supports broader participation, including geographically distributed and underrepresented communities, while Part B supports shared practices such as collaboration, knowledge sharing, and interoperability. Therefore, these two roadmap parts are foundational and deliver immediate benefits even when testbeds remain independent. We treat Part C as a long-term direction because global testbed connectivity requires sustained interoperability, shared interfaces across testbeds, and practical guidelines validated through multi-testbed experiments. 

The arrows in \Cref{fig_roadmap_parts_relation} summarize the interrelations between the three parts. Solid arrows indicate prerequisites for the long-term global testbed connectivity (Parts A and B enabling Part C) because global testbed connectivity benefits from a sufficiently broad community of participating testbeds (Part A) and from established best practices that reduce testbed-connecting efforts (Part B). Gray dashed arrows indicate reinforcing couplings (Parts A and B strengthening each other) and feedback from Part C back to Parts A and B (global testbed connectivity increasing the value of shared practices and participation). In particular, once multiple testbeds are connected, the benefits of shared tools, benchmarking, and communication interfaces become more visible, which can motivate further contributions and adoption.

\section{Conclusions}\label{sec:conclusions}
In this article, we proposed a roadmap to address the current challenges in small-scale testbeds for \acp{cav} and robot swarms. Drawing on discussions from our ``1\textsuperscript{st} Workshop on Small-Scale Testbeds for Connected and Automated Vehicles and Robot Swarms'' held at the IEEE IV 2024, we identified two main categories of challenges: transition challenges, specifically the \ac{sim2real} gap and small- to full-scale transition, and inherent real-world challenges, such as environmental unpredictability and the demands of distributed computing.

Our roadmap consists of three parts. Part A enhances accessibility and diversity to expand participation, especially for underrepresented communities. Part B shares best practices for the development, integration, and maintenance of testbeds, with an emphasis on collaboration and open-source contributions. Part C targets global testbed connectivity to support cross-site experimentation and collaboration. To support actionable planning, we summarized the traceability between the four challenges and the roadmap parts, and the interrelations between the roadmap parts.

\section*{Acknowledgement}
We thank Jinhyun Kim (Gyeongsang National University, South Korea) and Michele Magno (ETH Zurich, Switzerland) for their contributions to this work, as well as for their contributions as invited speakers to the workshop. We also thank Frank J. Jiang (KTH, Sweden), Kleio Fragkedaki (KTH, Sweden), and Lorenzo Lyons (Delft University of Technology, the Netherlands) for their contributions to this work, as well as for their contributions as authors of contributed papers to the workshop.

We acknowledge the financial support for this project by the Collaborative Research Center / Transregio 339 of the German Research Foundation (DFG).

\bibliographystyle{IEEEtran}
\bibliography{00_references}

\vskip -2\baselineskip plus -1fil
\begin{IEEEbiography}[{\includegraphics[width=1in,height=1.25in,clip,keepaspectratio]{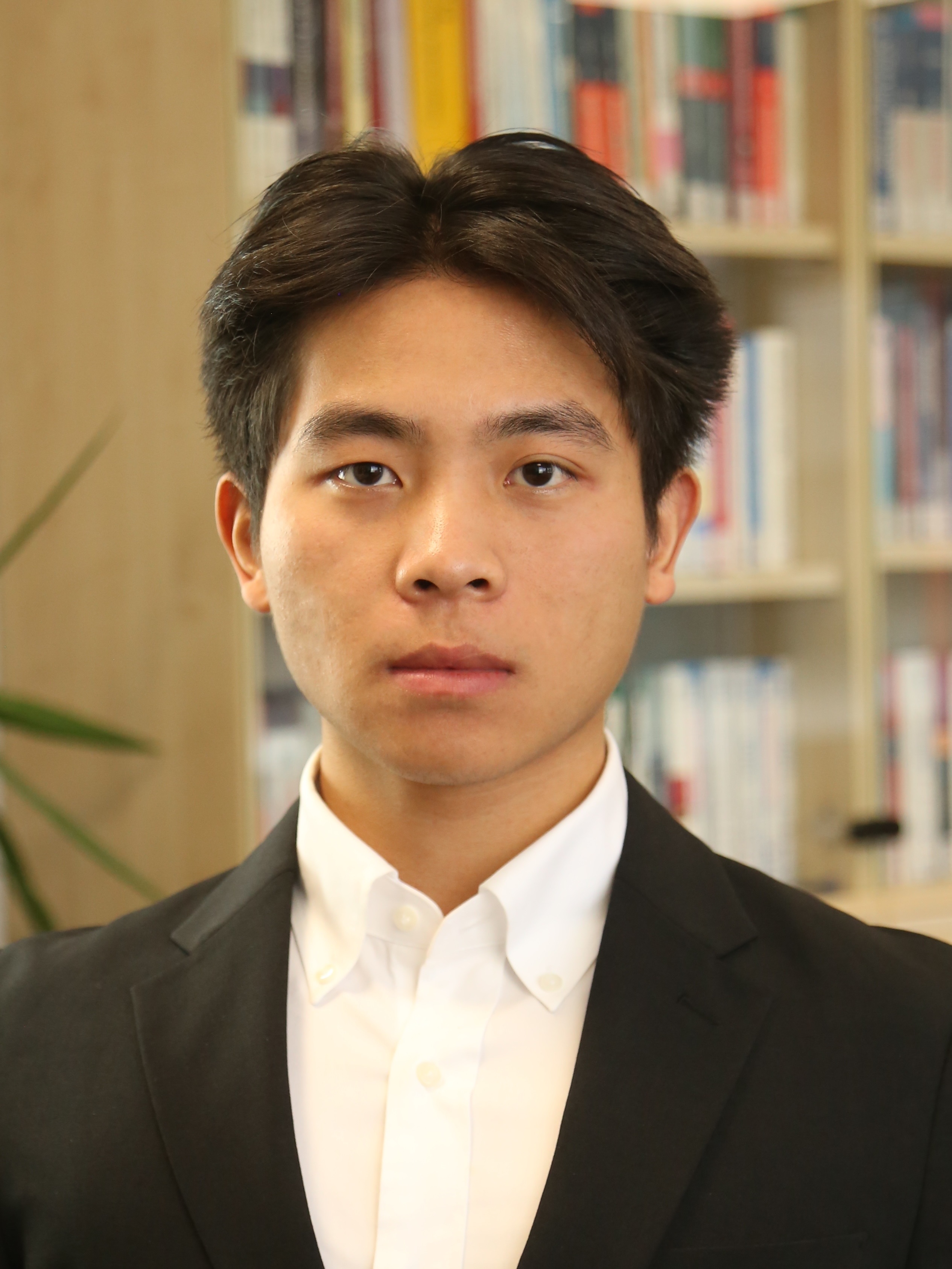}}]
{Jianye Xu} (Graduate Student Member, IEEE) Jianye Xu received the B.Sc. degree with distinction in Mechanical Engineering from the Beijing Institute of Technology, China, in 2020, and the M.Sc. degree with distinction in Automation Engineering from RWTH Aachen University, Germany, in 2022. He is currently pursuing a Ph.D. in Computer Science at RWTH Aachen University. His research focuses on learning- and optimization-based multi-agent decision-making and its applications in connected and automated vehicles.
\end{IEEEbiography}

\vskip -2\baselineskip plus -1fil
\begin{IEEEbiography}[{\includegraphics[width=1in,height=1.25in,clip,keepaspectratio]{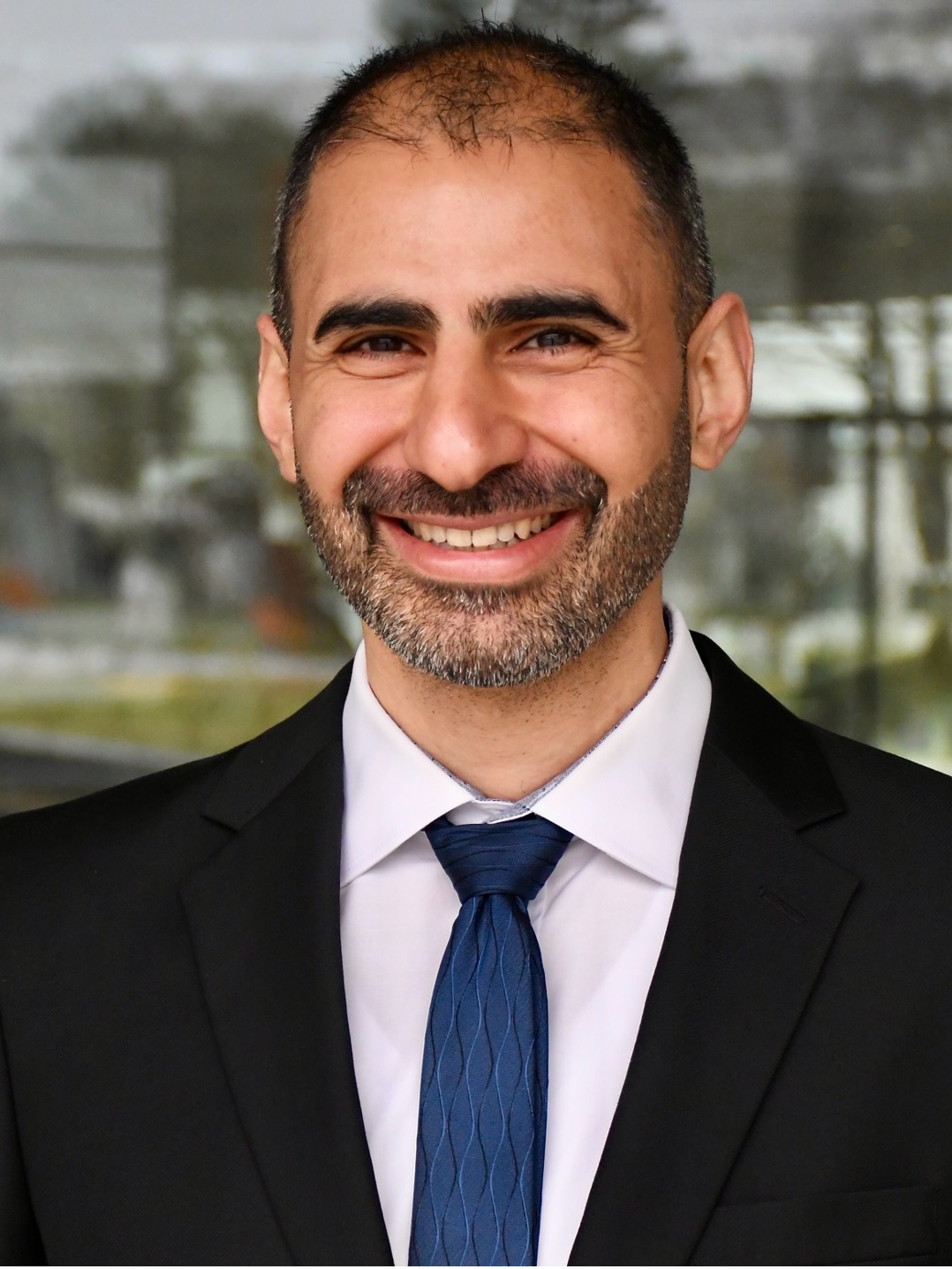}}]
{Bassam Alrifaee} (Senior Member, IEEE) Prof. Bassam Alrifaee holds the Professorship for Adaptive Behavior of Autonomous Vehicles in the Department of Aerospace Engineering at the University of the Bundeswehr (UniBw) Munich. His research focuses on the intelligent control of autonomous systems, with particular emphasis on distributed control, cooperative localization, software architectures, and experimental validation.
Before joining UniBw Munich in 2024, he served as a Senior Researcher and Lecturer at RWTH Aachen University, where he founded the Cyber-Physical Mobility (CPM) group and the CPM Lab. In 2023, he was a Visiting Scholar at the Information and Decision Science Laboratory at the University of Delaware, USA.
Prof. Alrifaee has secured research grants from various institutions and received awards for his advisory and editorial contributions. He is a Senior Member of the IEEE.
\end{IEEEbiography}

\vskip -2\baselineskip plus -1fil
\begin{IEEEbiography}[{\includegraphics[width=1in,height=1.25in,clip,keepaspectratio]{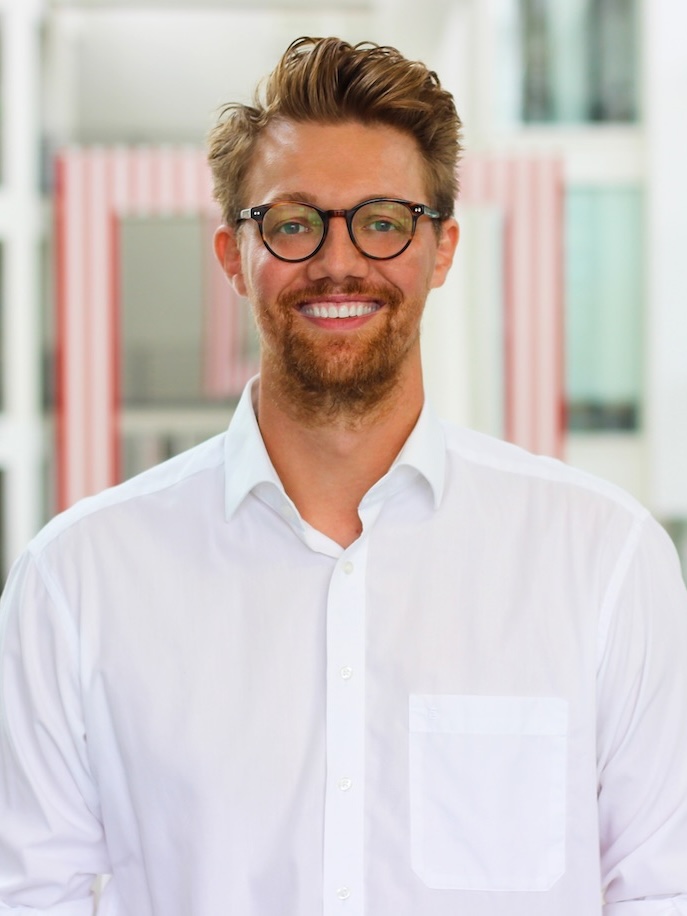}}]
{Johannes Betz} (Member, IEEE) Johannes Betz is an assistant professor in the Department of Mobility Systems Engineering at the Technical University of Munich (TUM), where he is leading the Autonomous Vehicle Systems (AVS) lab. He is one of the founders of the TUM Autonomous Motorsport team. His research focuses on developing adaptive dynamic path planning and control algorithms, decision-making algorithms that work under high uncertainty in multi-agent environments, and validating the algorithms on real-world robotic systems. Johannes earned a B.Eng. (2011) from the University of Applied Science Coburg, a M.Sc. (2012) from the University of Bayreuth, an M.A. (2021) in philosophy from TUM, and a Ph.D. (2019) from TUM. 
\end{IEEEbiography}

\vskip -2\baselineskip plus -1fil
\begin{IEEEbiography}[{\includegraphics[width=1in,height=1.25in,clip,keepaspectratio]{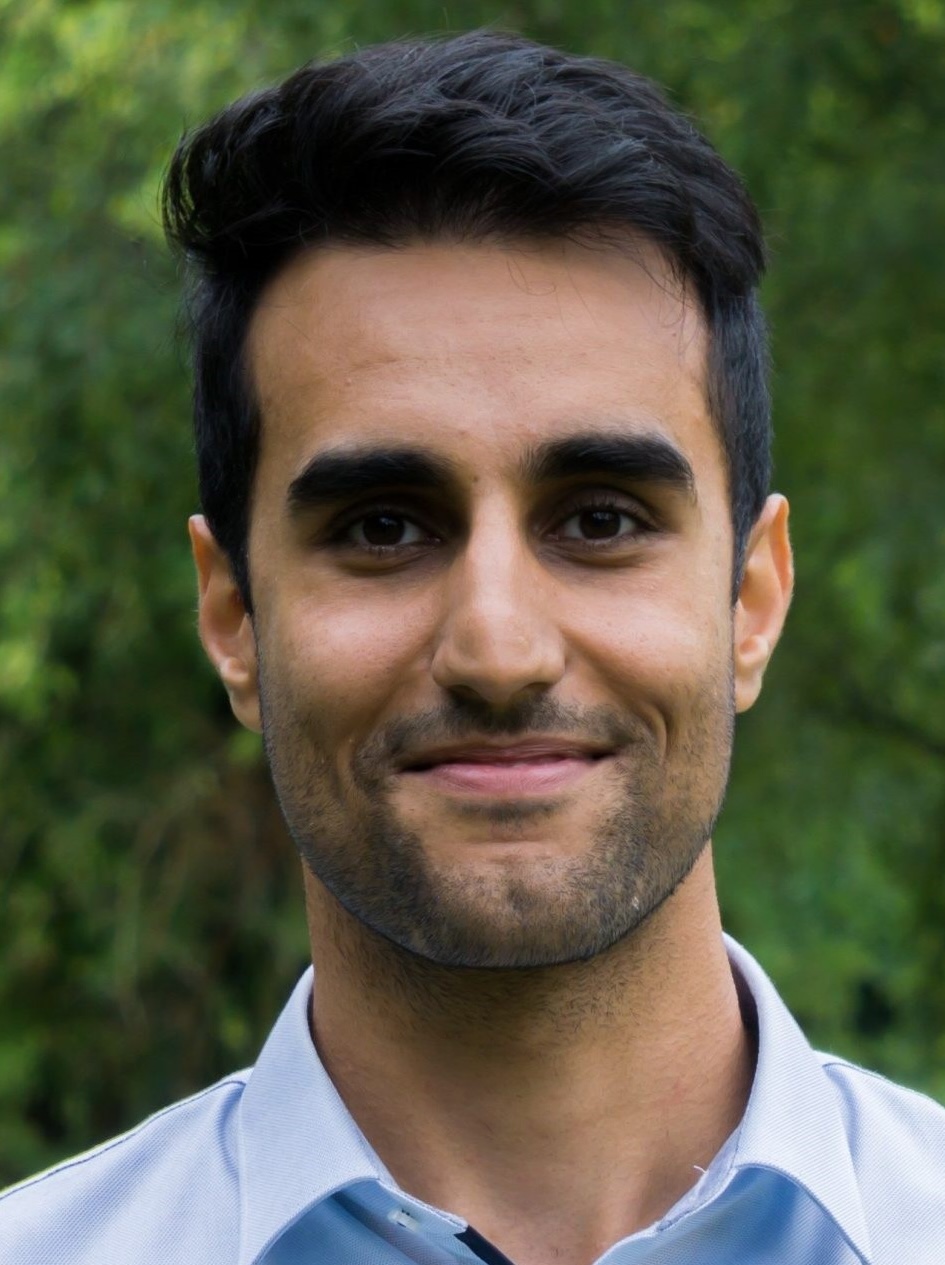}}]
{Armin Mokhtarian} completed his Ph.D. at RWTH Aachen University in 2024 after five years as a Research Assistant at the Chair of Embedded Software. His research focused on Connected and Automated Vehicles, exploring service-oriented software architectures and cloud-based solutions.
\end{IEEEbiography}

\vskip -2\baselineskip plus -1fil
\begin{IEEEbiography}[{\includegraphics[width=1in,height=1.25in,clip,keepaspectratio]{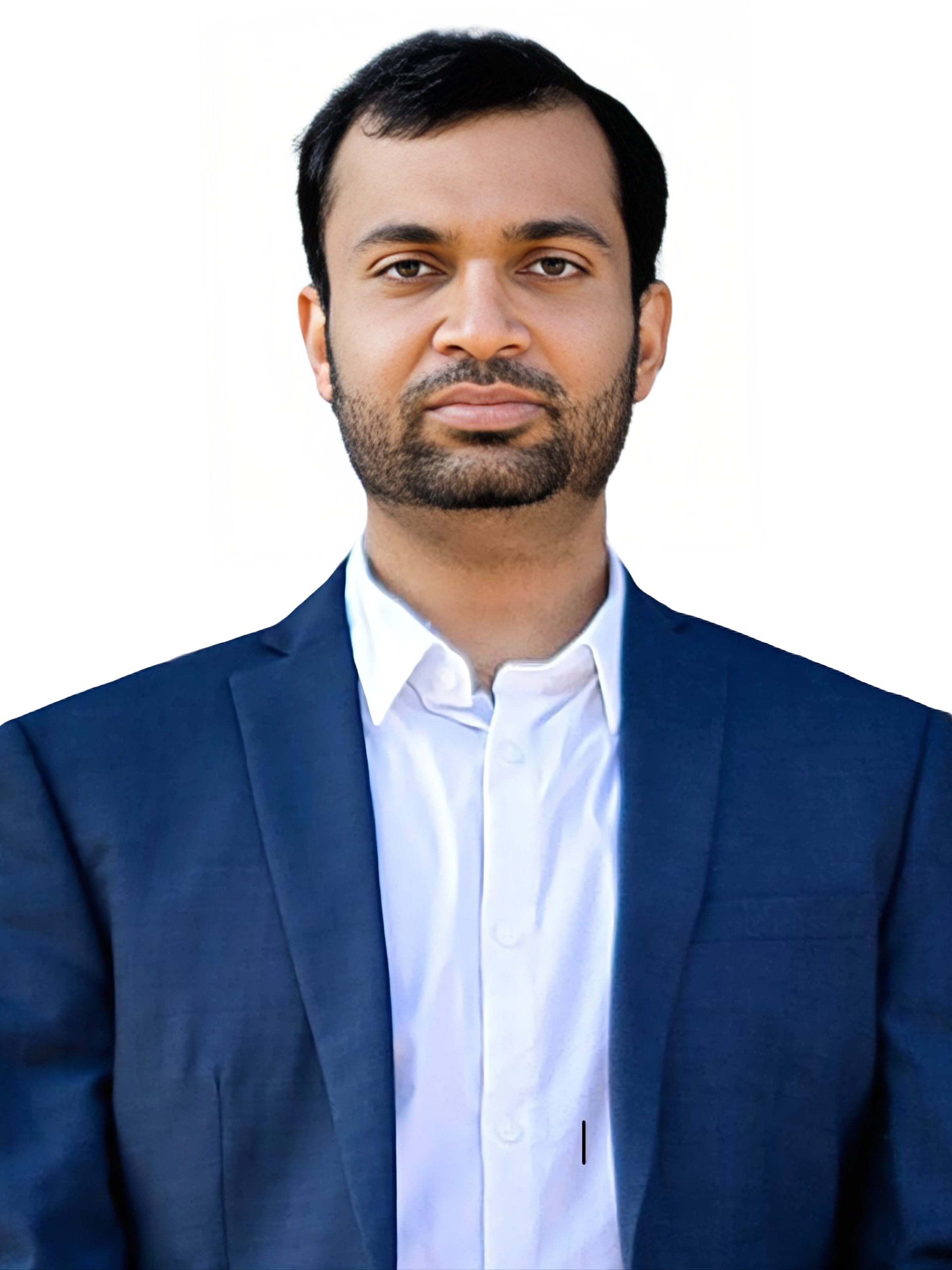}}]
{Archak Mittal} Dr. Archak Mittal, P.E., Ph.D., is an Assistant Professor in the Department of Civil Engineering at IIT Bombay. He holds a Ph.D. and M.S. in Civil and Environmental Engineering from Northwestern University, USA, and a B.Tech. with Honors from IIT Bombay. His expertise lies in intelligent transportation systems, traffic flow modeling, predictive analytics, and emerging vehicle technologies.
\end{IEEEbiography}

\vskip -2\baselineskip plus -1fil
\begin{IEEEbiography}[{\includegraphics[width=1in,height=1.25in,clip,keepaspectratio]{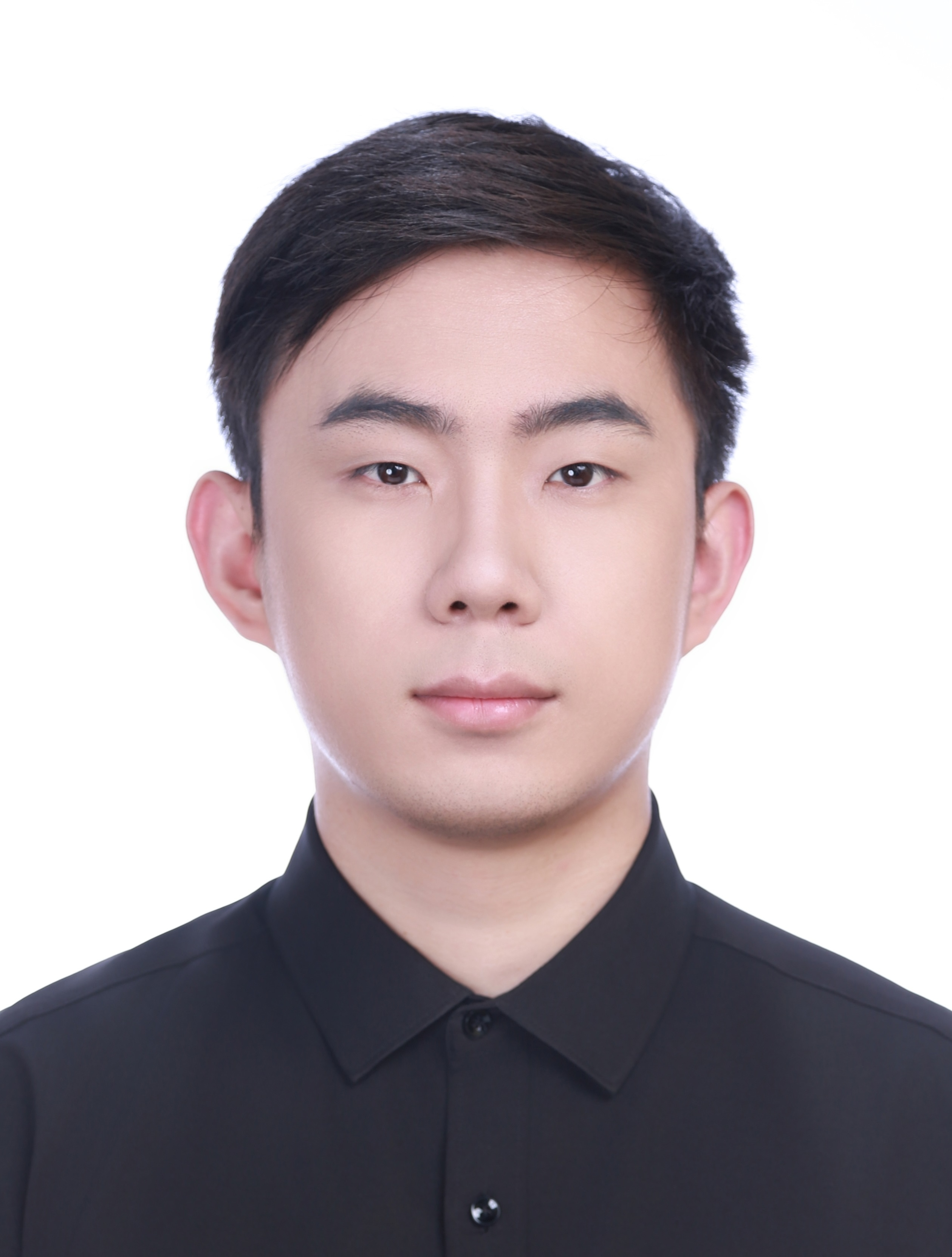}}]
{Mengchi Cai} received his B.E. degree and Ph. D. degree from School of Vehicle and Mobility, Tsinghua University, Beijing, China, in 2018 and 2023, respectively. He is currently a postdoctoral researcher at Intelligent and Connected Vehicles Lab, School of Vehicle and Mobility, Tsinghua University. Dr. Cai was recipients of the Young Elite Scientists Sponsorship Program by China Association for Science and Technology, the National Scholarship and Comprehensive Excellent Scholarship in Tsinghua University. He has been honored the Shuimu Tsinghua Scholarship, and led his team to win the first place in the OnSite Autonomous Driving Algorithm Challenge Competition. He is the author of over 30 research papers and over 20 patents, and he was the recipient of 1 best youth paper award of international conferences. His research interests include connected and automated vehicles and transportation, multi-vehicle formation control, and unsignalized intersection cooperation. 
\end{IEEEbiography}

\vskip -2\baselineskip plus -1fil
\begin{IEEEbiography}[{\includegraphics[width=1in,height=1.25in,clip,keepaspectratio]{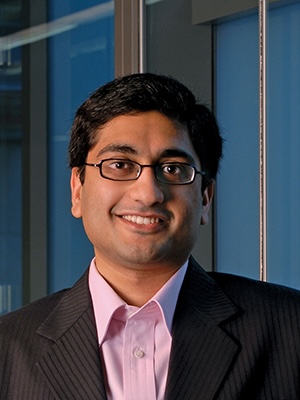}}]
{Rahul Mangharam} (Senior Member, IEEE) Professor of Electrical \& Systems Engineering, University of Pennsylvania. He leads xLAB for building Safe Autonomous Systems and conducts research at the intersection of formal methods, machine learning and controls. He is the Penn Director for the US DoT's \$20MM Safety21 National University Transportation Center. Rahul received the 2016 US Presidential Early Career Award (PECASE) from President Obama and has won several ACM and IEEE best paper awards. He founded RoboRacer.AI for teaching the foundations of autonomous vehicles across 90+ universities worldwide. 
\end{IEEEbiography}

\vskip -2\baselineskip plus -1fil
\begin{IEEEbiography}[{\includegraphics[width=1in,height=1.25in,clip,keepaspectratio]{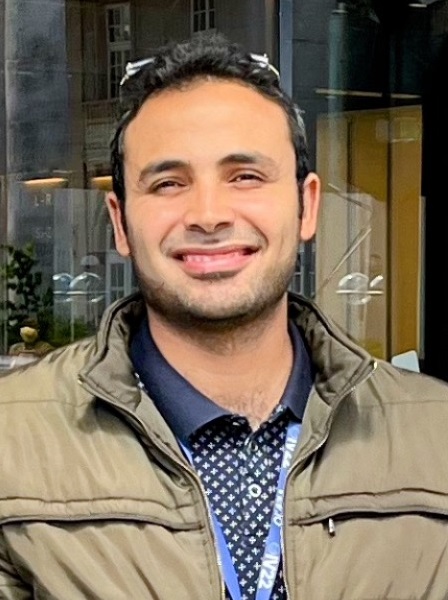}}]
{Omar M. Shehata} (Member, IEEE) is an Associate Professor in the Mechatronics Engineering department, German University in Cairo (GUC), Egypt. He founded and directs the Multi-Robot Systems (MRS) research group since 2015, with a special attention to the utilization of scaled-vehicles in the validation of \ac{its} applications and algorithms. His research aims at bridging the research worldwide and the unique challenges of developing regions focusing on cooperative control of \acp{cav}, optimization algorithms and intelligent control applications.
\end{IEEEbiography}

\vskip -2\baselineskip plus -1fil
\begin{IEEEbiography}[{\includegraphics[width=1in,height=1.25in,clip,keepaspectratio]{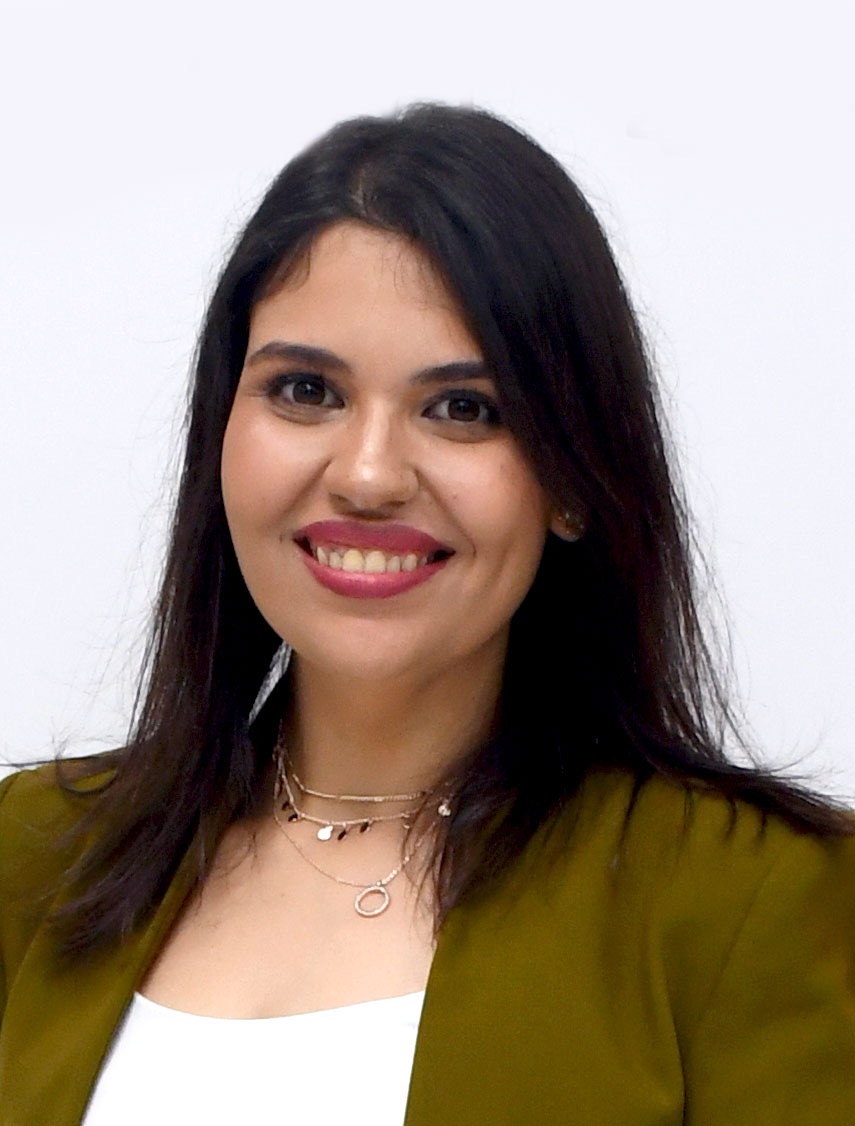}}]
{Catherine M. Elias} Dr. Eng. Catherine M. Elias is a lecturer in the Computer Science and Engineering Department, Faculty of Media Engineering and Technology (MET). She received her Ph.D. degree in Mechatronics Engineering from the GUC in Dec. 2022 in the field of Cooperative Architecture for Transportation Systems. She is currently the director of the Cognitive Driving Research in Vehicular Systems (C-DRiVeS Lab), working in the field of autonomous driving stack modules. She serves as a Board of Governors (BoG) member in the IEEE \ac{its} Society during the interval 2023-2025, the 2023-2025 Co-chair of the committee on Diversity, Equity, and Inclusion in \ac{its} committee chair.
\end{IEEEbiography}

\vskip -2\baselineskip plus -1fil
\begin{IEEEbiography}[{\includegraphics[width=1in,height=1.25in,clip,keepaspectratio]{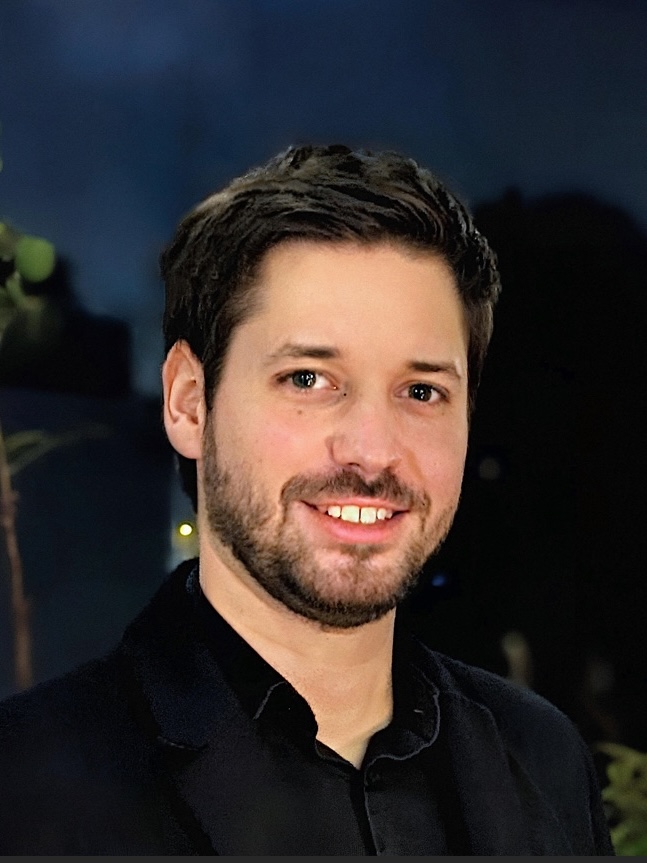}}]
{Jan-Nico Zaech} Dr. Jan-Nico Zaech, research scientist at INSAIT, Sofia University ``Sofia University St. Kliment Ohridski'', leads a research group on robotics and embodied AI. He received his Ph.D. at ETH Zurich with work on autonomous driving and quantum computer vision. His research revolves around robotic foundation models, with a focus on the integration of 3D data, visual robustness, and modular architectures. 
\end{IEEEbiography}

\vskip -2\baselineskip plus -1fil
\begin{IEEEbiography}[{\includegraphics[width=1in,height=1.25in,clip,keepaspectratio]{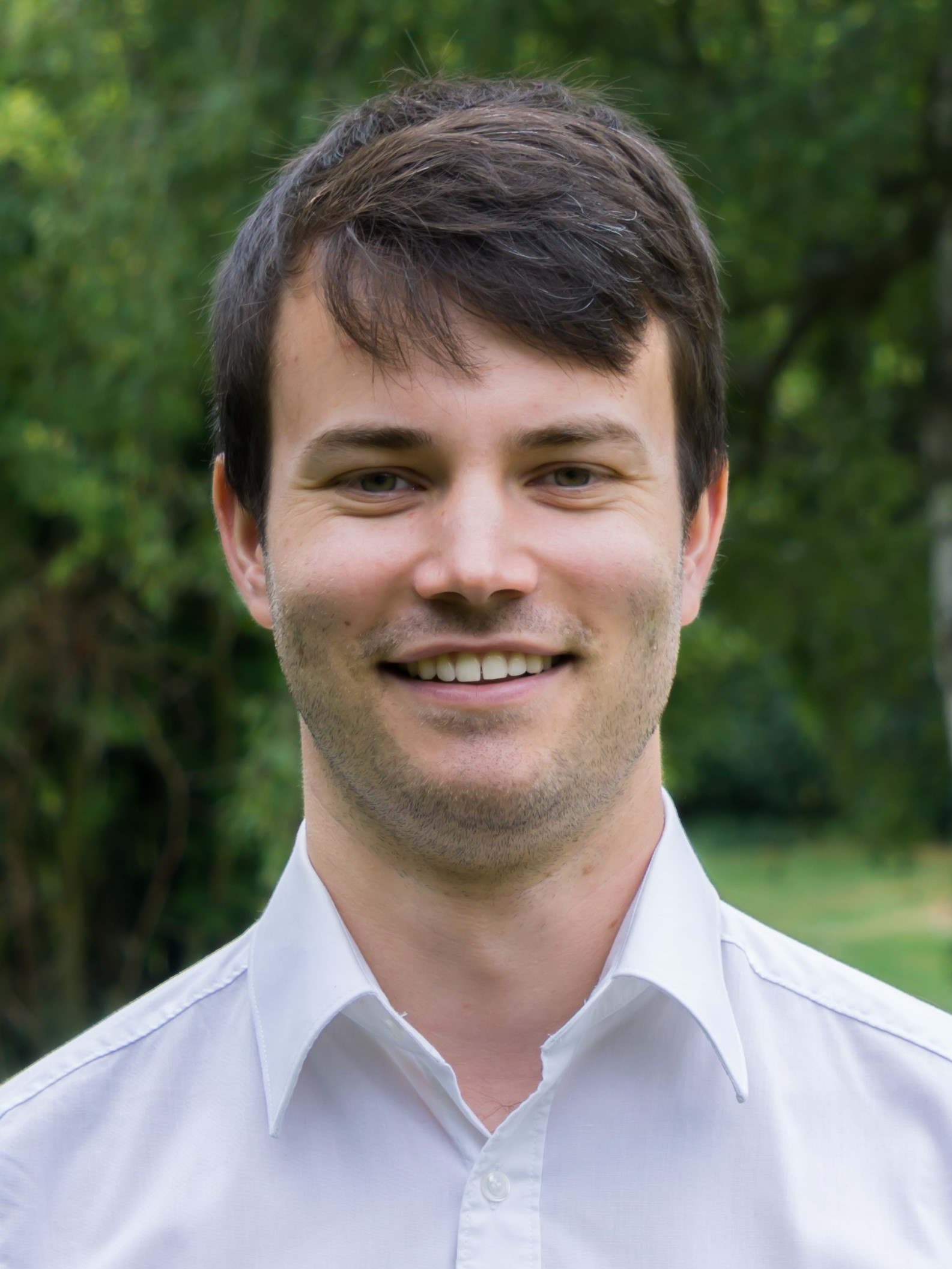}}]
{Patrick Scheffe} (Graduate Student Member, IEEE)
is a postdoctoral researcher at KU Leuven, Belgium.
He has obtained a Ph.D. degree in Engineering in 2025 from RWTH Aachen University, Germany.
His research revolves around multi-agent decision-making and its application to motion planning for connected and autonomous vehicles, both on-road and off-road.
\end{IEEEbiography}

\vskip -2\baselineskip plus -1fil
\begin{IEEEbiography}[{\includegraphics[width=1in,height=1.25in,clip,keepaspectratio]{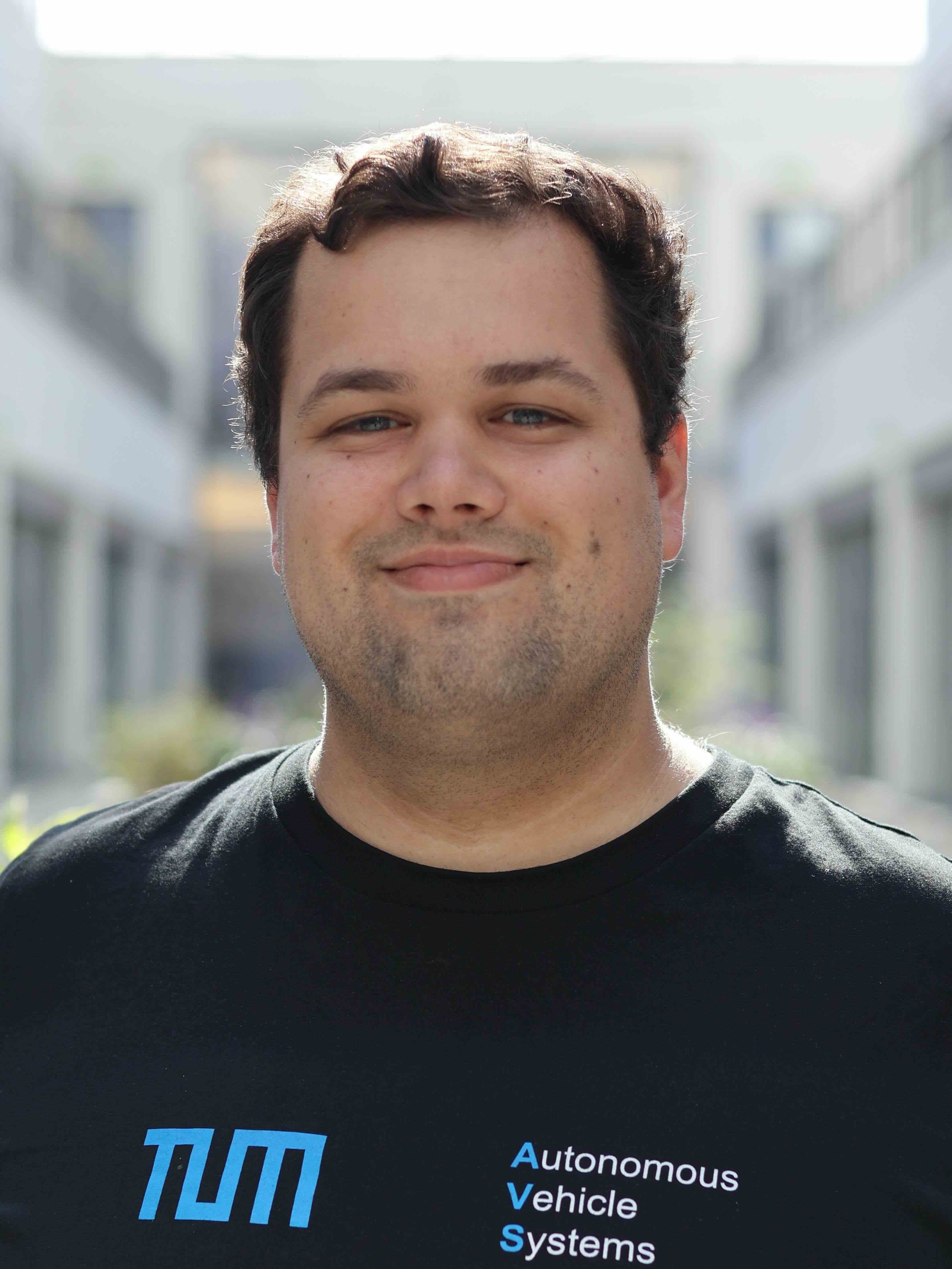}}]
{Felix Jahncke} received his bachelor's degree in Mechanical Engineering and master's degree in Automotive Engineering from the Technical University of Munich. During his master's studies, he completed research visits at the Technical University of Delft (Netherlands) and the University of Pennsylvania (USA). Since 2023, he has been pursuing his Ph.D. at TUM, focusing on the transition from traditional software architectures towards end-to-end learning methods, using F1TENTH vehicles.
\end{IEEEbiography}

\vskip -2\baselineskip plus -1fil
\begin{IEEEbiography}[{\includegraphics[width=1in,height=1.25in,clip,keepaspectratio]{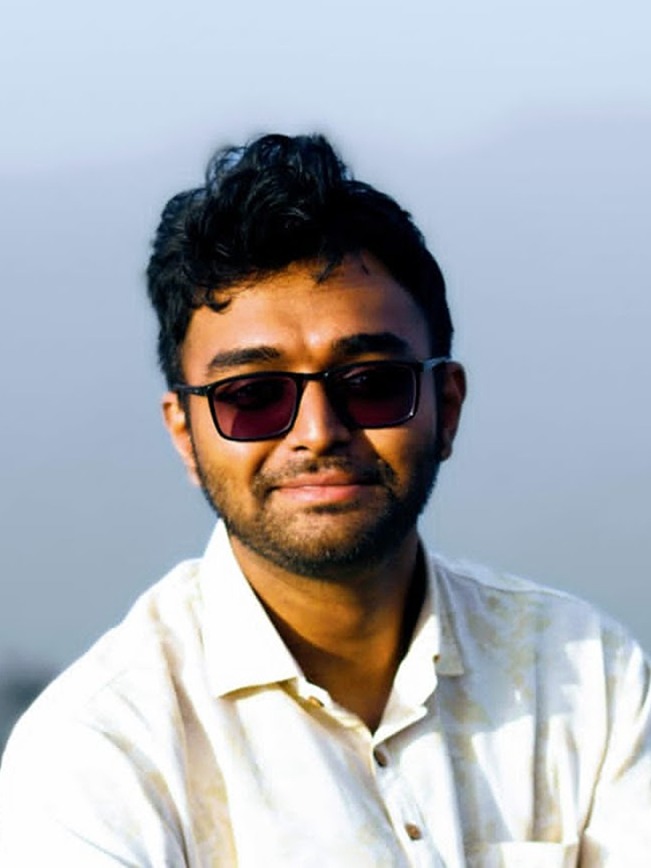}}]
{Sangeet S. Ulhas} received his B.Tech. in Naval Architecture from CUSAT (India) in 2017 and an M.S. in Mechanical Engineering from Arizona State University (USA) where he is now pursuing his Ph.D. in Mechanical Engineering, affiliated to the Autonomous Collective Systems (ACS) Laboratory. His research focuses on Ethical Decision Making and Control of AVs, Optimization, CV and Sim2Real transfer.
\end{IEEEbiography}

\vskip -2\baselineskip plus -1fil
\begin{IEEEbiography}[{\includegraphics[width=1in,height=1.25in,clip,keepaspectratio]{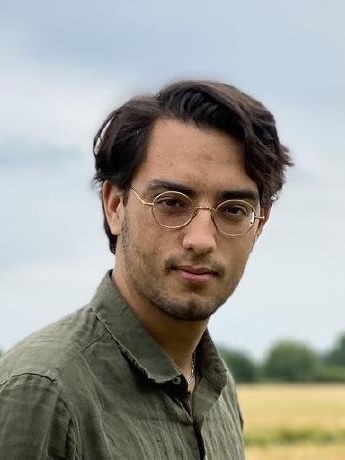}}]
{Kaj Munhoz Arfvidsson} is a doctoral student with the School of Electrical Engineering and Computer Science at KTH Royal Institute of Technology in Stockholm, Sweden. He is also affiliated with the Integrated Transport Research Lab (ITRL). He received his degree of Master of Science in Engineering, with specialization in Systems, Control and Robotics, from KTH Royal Institute of Technology in 2025. His research interests include formal verification, control, vehicular connectivity and automation, and intelligent transportation systems.
\end{IEEEbiography}

\end{document}